\documentclass[
]{ceurart}

\usepackage{listings}
\usepackage{amssymb}
\usepackage{amsmath}
\usepackage{caption} 
\usepackage{comment}

\usepackage{booktabs}
\usepackage{makecell}
\usepackage{tabularx}
\usepackage{array}

\usepackage[table,xcdraw]{xcolor}

\usepackage{acronym}
\acrodef{BioNLP}[BioNLP]{Biomedical Natural Language Processing}
\acrodef{BIO}[BIO]{Beginning-Inside-Outside}
\acrodef{C-RE}[C-RE]{Concept-level RE}
\acrodef{DocRE}[DocRE]{Document-Level Relation Extraction}
\acrodef{EFCL}[EFCL]{Extract First Classify Later}
\acrodef{IAA}[IAA]{Inter-Annotator Agreement}
\acrodef{IE}[IE]{Information Extraction}
\acrodef{KB}[KB]{Knowledge Base}
\acrodef{KOS}[KOS]{Knowledge Organization System}
\acrodef{HPC}[HPC]{High Performance Computing}
\acrodef{IR}[IR]{Information Retrieval}
\acrodef{LM}[LM]{Language Model}
\acrodef{LLM}[LLM]{Large Language Model}
\acrodef{M-RE}[M-RE]{Mention-level RE}
\acrodef{NER}[NER]{Named Entity Recognition}
\acrodef{NERD}[NERD]{Named Entity Recognition and Disambiguation}
\acrodef{NEL}[NEL]{Named Entity Linking}
\acrodef{NLP}[NLP]{Natural Language Processing}
\acrodef{OIE}[OIE]{Open Information Extraction}
\acrodef{RE}[RE]{Relation Extraction}
\acrodef{SOTA}[SOTA]{State Of The Art}
\acrodef{TC}[TC]{Term Classification}
\acrodef{TE}[TE]{Term Extraction}
\acrodef{TWIX}[TWIX]{\textbf{T}wo-stage \textbf{W}orkflow for \textbf{I}nformation e\textbf{X}traction}
\acrodef{UI}[UI]{User Interface}
\acrodef{URI}[URI]{Uniform Resource Identifier}

\begin{document}

\copyrightyear{2026}
\copyrightclause{Copyright for this paper by its authors.
  Use permitted under Creative Commons License Attribution 4.0
  International (CC BY 4.0).}

\conference{CLEF 2026 Working Notes, 21 -- 24 September 2026, Jena, Germany}

\title{TWIX: a Two-Stage Approach for End-To-End Named Entity Recognition and Relation Extraction}

\author[1]{Marco Martinelli}[%
orcid=0009-0001-1596-8642,
email=martinell2@dei.unipd.it,
url=https://www.dei.unipd.it/~martinell2/,
]
\cormark[1]
\fnmark[1]

\author[1]{Laura Menotti}[%
orcid=0000-0002-0676-682X,
email=laura.menotti@unipd.it,
url=https://www.dei.unipd.it/~menottilau/,
]
\cormark[1]
\fnmark[1]

\address[1]{Department of Information Engineering, University of Padova, Padova, Italy}

\cortext[1]{Corresponding authors.}
\fntext[1]{These authors contributed equally.}

\begin{abstract}
  The exponential growth of scientific publications calls for automatic \ac{IE} systems to support knowledge discovery. In this context, the GutBrainIE benchmark evaluates \ac{NER}, \ac{NERD}, and \ac{RE} systems in the gut-brain axis domain. We propose \ac{TWIX}, an end-to-end \ac{IE} pipeline featuring three interconnected modules, each leveraging a two-stage framework to solve all four GutBrainIE subtasks. 
  Evaluation on the development and test sets shows that our method substantially outperforms the baseline by a wide margin, while also ranking first among all participant submissions across all subtasks.
  These results indicate that the proposed two-stage pipeline effectively improves both precision and recall in practical settings.
\end{abstract}


\maketitle

\section{Introduction}
\label{sec:introduction}

\begin{figure}[t]
    \centering
    \includegraphics[width=0.8\linewidth]{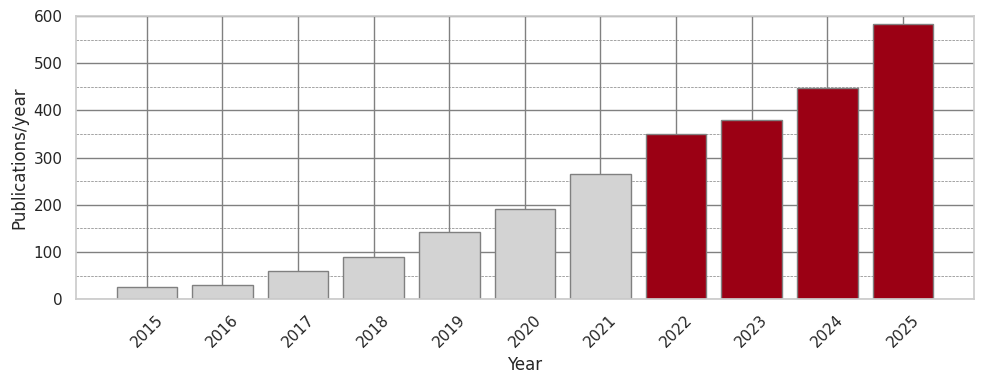}
    \caption{Annual number of PubMed publications related to the gut--brain axis, retrieved on May 27, 2026 using the query adopted to populate the GutBrainIE 2026 collection: \textit{((gut) AND (microbiota OR microbiome)) AND (dementia OR Alzheimer OR multiple sclerosis OR Amyotrophic lateral sclerosis)}. Red bars highlight years with more than 300 records.}
    \label{fig:PubMed_Timeline_Results_by_Year}
\end{figure}

Recent biomedical research increasingly links the gut microbiota to neurological and psychiatric diseases and disorders, including Parkinson's disease, Alzheimer's disease, Multiple Sclerosis, mood disorders, and mental health conditions more broadly~\cite{carabotti2015gut,ghaisas2016gut,appleton2018gut,cryan2020gut}. This growing interest is reflected in the rapid expansion of the scientific literature on the gut-brain axis. As shown in Figure~\ref{fig:PubMed_Timeline_Results_by_Year}, the number of PubMed publications in this area more than doubled between 2020 and 2025, increasing from around 300 to more than 700 articles per year. While this trend highlights the relevance of the field, it also creates a substantial challenge for clinicians and researchers, who must identify, interpret, and connect relevant findings across a rapidly growing body of unstructured scientific texts~\cite{martinelli2026domain}.

Automatic \ac{IE} systems can support knowledge discovery in this setting by transforming unstructured biomedical text into structured information. In this context, the GutBrainIE challenge provides a benchmark of PubMed abstracts focused on the gut-brain axis and its implications for Parkinson's disease, Alzheimer's disease, Multiple Sclerosis, and mental health. The dataset includes annotations for entity mentions, concept-level links, and semantic relations, and defines four subtasks of increasing complexity: \ac{NER}, requiring systems to detect and classify biomedical entity mentions; \ac{NERD}, extending \ac{NER} by linking each mention to a concept from biomedical reference vocabularies; \ac{M-RE}, requiring the extraction of semantic relations between entity mentions; and \ac{C-RE}, which abstracts \ac{RE} from surface mentions to linked biomedical concepts.

All subtasks are framed in an end-to-end setting, where the test documents are released without ground-truth annotations. As a consequence, when approached with modular \ac{IE} pipelines, downstream tasks depend on the predictions produced by upstream components. For instance, \ac{M-RE} requires entity mentions predicted by an upstream \ac{NER} module, while \ac{C-RE} requires linked entity mentions produced by an upstream \ac{NERD} module. This setting makes error propagation a central issue: false positives or incorrect predictions in earlier stages can directly affect the performance of subsequent modules.

To address this challenge, we propose \acf{TWIX}, a precision-oriented \ac{IE} pipeline composed of three interconnected modules, each following a two-stage strategy. The \ac{NER} module adopts an \ac{EFCL} architecture, where candidate mentions are first extracted without assigning entity labels and then validated and classified by a separate component. This design introduces two filtering steps, reducing false positive mentions before they are passed to downstream modules. The \ac{NEL} module follows a retrieve-and-rerank strategy: a retriever first selects the top-$k$ candidate concepts for each mention, and a reranker then selects the most likely concept in context. Finally, the \ac{RE} module builds on ATLOP, the architecture used in the official baseline, but modifies the training procedure by first pre-training on larger, lower-quality data and then fine-tuning on smaller, higher-quality annotations~\cite{atlop}.

Experimental results on the development set show that our architecture substantially outperforms standard baselines and competitive approaches inspired by top-performing systems from the 2025 edition of GutBrainIE~\cite{martinelli2025overview}. Results on the test set further confirm the effectiveness of \ac{TWIX}, with our system ranking first across all four subtasks. Moreover, the performance gap with respect to the second-best system increases across the subtasks, suggesting that the precision-oriented design is particularly beneficial in reducing error propagation in end-to-end \ac{IE} settings. We share the codebase in GitHub at: \url{https://github.com/MMartinelli-hub/GBIE26_TWIX/}.

The remainder of the paper is organized as follows. Section~\ref{sec:related-work} discusses the related work guiding the design of \ac{TWIX}. Section~\ref{sec:data-task-description} describes the GutBrainIE data and subtasks. Section~\ref{sec:methodology} presents the proposed architecture and its modules. Section~\ref{sec:experiments-and-results} reports the development-set experiments and component-level analyses. Section~\ref{sec:submission-results} presents and discusses the official test-set results. Finally, Section~\ref{sec:conclusion} concludes the paper and outlines future research directions.

\section{Related Work}
\label{sec:related-work}

End-to-end \ac{NER} and \ac{RE} are commonly addressed through either pipeline or joint architectures. 
Pipeline approaches decompose the task into separate stages, first identifying entity mentions and then predicting relations over candidate entity pairs derived from the \ac{NER} output~\cite{zhong2021frustratingly,yan2022empirical}. Joint and multi-task models, instead, model entities and relations simultaneously, with the goal of reducing error propagation between subtasks~\cite{bekoulis2018joint,shang2022onerel,stepanov2026gliner}. 
Although pipeline architectures may suffer from propagation errors, recent work has shown that simple and well-engineered pipelines remain highly competitive, while also making it easier to analyze the contribution and limitations of individual components~\cite{zhong2021frustratingly}. 
Following this line, we adopt a modular pipeline design, which allows us to explicitly study and control error propagation across \ac{NER}, \ac{NEL}, and \ac{RE} modules.

For \ac{NER}, the most common approach formulates the task as token classification, typically using a pretrained transformer encoder followed by a token-level classification head~\cite{vacareanu2024active}. This was also the dominant strategy in the 2025 edition of the GutBrainIE challenge~\cite{martinelli2025overview}. However, several works have shown that extending this standard formulation can improve effectiveness, for example by incorporating external resources or by introducing auxiliary training objectives~\cite{tong2021multi,fu2023biomedical}. 
More recently,~\cite{shlyk-etal-2026-mind} proposed a two-stage \ac{NER} strategy in which entities are first extracted and then tagged. While their approach is implemented through \ac{LLM}-based prompting, we adapt the same general principle to supervised biomedical transformer models, formulating entity extraction as token classification and entity tagging as sequence classification.

For \ac{NEL}, similarity-based architectures are widely adopted. These methods represent mentions and candidate concepts in a shared space and select the concept most similar to the mention in context~\cite{logeswaran-etal-2019-zero,wu-etal-2020-scalable}. A common design follows a retrieve-and-rerank strategy inspired by \ac{IR}: a retriever first selects a ranked top-$k$ subset of candidate concepts from the full vocabulary, and a reranker then performs a more accurate contextual disambiguation over this reduced candidate set~\cite{nogueira2019passage}. Dual encoders are commonly used for efficient retrieval, while cross encoders are often employed for reranking due to their ability to jointly model the mention context and candidate concept description~\cite{wu-etal-2020-scalable,logeswaran-etal-2019-zero}. In this work, we follow this two-stage \ac{NEL} paradigm by combining a retriever and a reranker. 

State-of-the-Art \ac{DocRE} models feature sequence-based approaches that treat documents as a sequence of tokens and learn contextual representations for all entity pairs. Early approaches using CNNs and LSTMs struggled with long-range dependencies~\cite{yao_etal-2019}. Transformer-based models leveraging contextual embeddings, such as BERT, markedly improve the task and are now standard in \ac{DocRE}~\cite{devlin2019bert}. A key transformer-based model, which is also the baseline of the Relation Extraction subtasks, is ATLOP~\cite{atlop}. ATLOP introduces an adaptive threshold for entity-dependent multi-label classification and localized context pooling to improve the prediction of precise triples.

\begin{table}[t]
\centering
\caption{Statistics of the GutBrainIE collections. For each collection, the table reports the number of documents, the total number of annotated entities, the average number of entities per document, the total number of annotated relations, and the average number of relations per document.}
\label{tab:dataset-statistics}
\begin{tabular}{lrrrrr}
\toprule
Collection & \# Docs & \# Entities & Ents/Doc & \# Relations & Rels/Doc \\
\midrule
Train Gold        & 639  & 20,530 & 32.13 & 8,556  & 13.39 \\
Train Silver      & 811  & 26,134 & 32.22 & 10,907 & 13.45 \\
Train Silver 2025 & 499  & 15,275 & 30.61 & 10,616 & 21.27 \\
Train Bronze      & 2,972 & 89,987 & 30.28 & 29,692 & 9.99 \\
Development Set   & 80   & 2,521  & 31.51 & 1,261  & 15.76 \\
\bottomrule
\end{tabular}
\end{table}

\section{Data and Task Description}
\label{sec:data-task-description}

The GutBrainIE challenge is based on a collection of titles and abstracts of biomedical articles retrieved from PubMed, focusing on the gut-brain interplay and its implications for neurological and mental diseases.
The dataset provides annotations for entity mentions, concept-level links, and relations, supporting the development and evaluation of \ac{IE} systems able to extract structured biomedical knowledge from unstructured scientific texts. 

The corpus covers 13 entity types, including general biomedical categories, such as \textit{anatomical location}, \textit{bacteria}, \textit{chemical}, and \textit{drug}, as well as categories more specific to the gut-brain axis, such as \textit{microbiome} and \textit{dietary supplement}. The schema also includes entities related to experimental settings, such as \textit{biomedical technique} and \textit{statistical technique}. Entity mentions are linked to concept identifiers from six standardized biomedical vocabularies, together with a custom GutBrainIE ontology used for mentions that cannot be mapped to existing resources~\cite{martinelli2026domain}.

Relations are defined between annotated entities and are associated with one of 17 relation predicates. Since several predicates can connect different combinations of entity types, and the same pair of entity types may be associated with different predicates, the schema defines 55 distinct relation triples, corresponding to combinations of subject label, predicate, and object label.

The released data are organized into collections reflecting different levels of annotation quality. The \textit{Gold} collection contains expert-curated annotations. The \textit{Silver} collections contain annotations produced by trained students under expert supervision, including both the current Silver collection and the Silver 2025 collection from the previous edition, for which concept-level annotations were automatically generated. The \textit{Bronze} collection contains fully automatic annotations produced using fine-tuned GLiNER for \ac{NER} and fine-tuned ATLOP for \ac{RE}~\cite{gliner,atlop}. Table~\ref{tab:dataset-statistics} reports the main statistics for each collection. 

The challenge proposes four subtasks, grouped into two main families: entity extraction and normalization, and relation extraction~\cite{BioASQ2026,BioASQ2026taskGutBrainIE}. In Subtask 6.1.1 (\ac{NER}), participants are given PubMed titles and abstracts and are asked to identify entity mentions and classify them into one of the 13 predefined categories. Each entity mention is represented through its category, location in the document, and character offsets:
\[
(\textit{entityCategory}\,;\ \textit{entityLocation}\,;\ \textit{startOffset}\,;\ \textit{endOffset}).
\]

Subtask 6.1.2 (\ac{NERD}) extends \ac{NER} by requiring each predicted entity mention to be linked to a concept identifier from the GutBrainIE reference resources. Predictions therefore include the mention information together with the corresponding concept URI:
\[
(\textit{entityCategory}\,;\ \textit{entityLocation}\,;\ \textit{startOffset}\,;\ \textit{endOffset}\,;\ \textit{conceptURI}).
\]

In Subtask 6.2.1 (\ac{M-RE}), participants are asked to identify relations between specific entity mentions within a document. Each prediction specifies the subject mention, relation predicate, and object mention:
\[
(\textit{subjectMention}\,;\ \textit{relationPredicate}\,;\ \textit{objectMention}).
\]

Finally, Subtask 6.2.2 (\ac{C-RE}) abstracts \ac{RE} from surface mentions to linked biomedical concepts. Each prediction specifies the subject concept, its category, the relation predicate, the object concept, and its category:
\[
(\textit{subjectConceptURI}\,;\ \textit{subjectCategory}\,;\ \textit{relationPredicate}\,;\ \textit{objectConceptURI}\,;\ \textit{objectCategory}).
\]

For each subtask, the test set contains only the PubMed identifier, title, and abstract of each document, and participants are required to produce predictions in the format defined for the corresponding subtask.

All submitted runs are evaluated using Precision ($P$), Recall ($R$), and $F_1$ score, computed with both macro- and micro-averaging. For the \ac{NER} and \ac{NERD} subtasks, labels correspond to the 13 entity categories, while for the two \ac{RE} subtasks they correspond to relation triples defined by subject label, predicate, and object label. The official leaderboard uses micro-averaged $F_1$ as the reference metric, since it better accounts for the strong class imbalance characterizing both entity labels and relation predicates.

\section{Methodology}
\label{sec:methodology}
\begin{figure}[t]
    \centering
    \includegraphics[width=\linewidth]{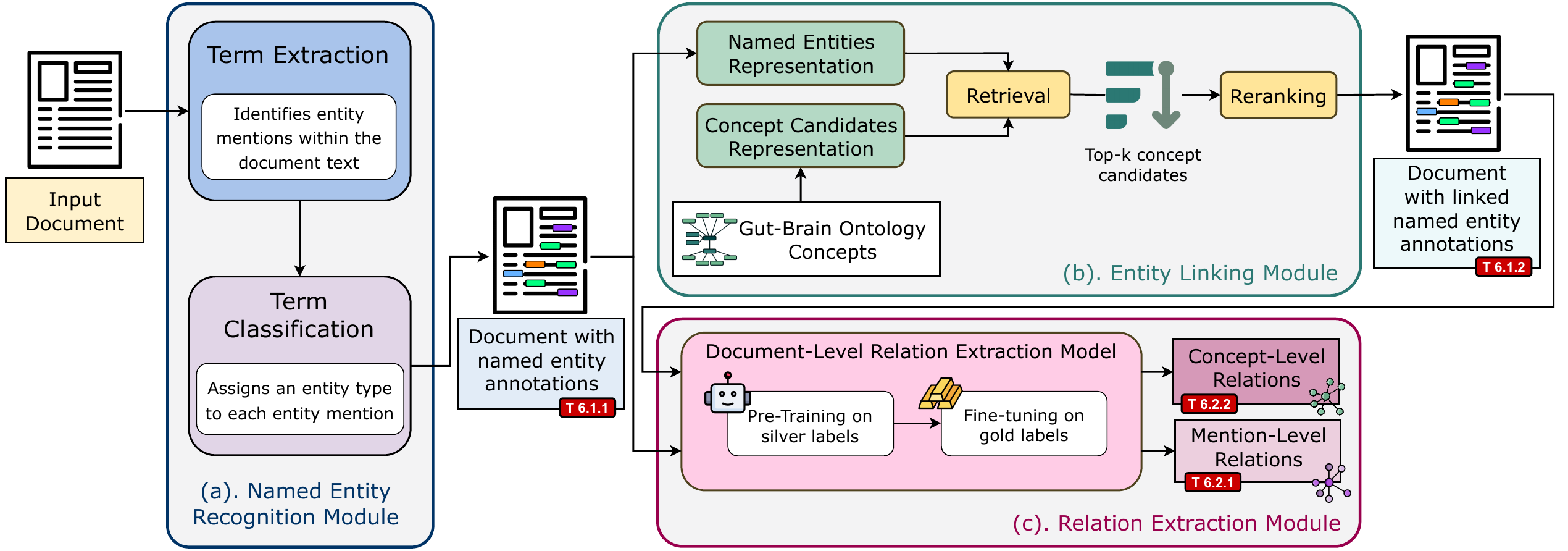}
    \caption{The \ac{TWIX} architecture including: (a) Named Entity Recognition module; (b) Entity Linking Module; (c) Relation Extraction module.}
    \label{fig:pipelineOverview}
\end{figure}

Figure~\ref{fig:pipelineOverview} presents \ac{TWIX}, reporting its main components and modules involved. The pipeline is composed of three interconnected modules, addressing \ac{NER}, \ac{NEL}, and \ac{RE}, respectively.
The first module performs \ac{NER} over the titles and abstracts of GutBrainIE documents. We address \ac{NER} through an \ac{EFCL} architecture, a two-stage design in which a \ac{TE} component first identifies candidate entity mentions in the input text, and a \ac{TC} component then validates these candidates and assigns one of the GutBrainIE entity labels to each of them.
The second module addresses \ac{NEL}. It receives the documents together with the entity mentions predicted by the \ac{NER} module and links each mention to a concept from the GutBrainIE reference resources. This module also follows a two-stage strategy: first, a retriever selects a set of candidate concepts for each mention; then, a reranker scores these candidates in context and selects the most likely concept as the final prediction.
The third module performs \ac{RE} over the annotated documents, featuring a two-stage training built upon ATLOP. In particular, the \ac{RE} module is pre-trained for a few epochs on noisy labels and then finetuned on high-quality annotations.

\subsection{Selection of Pretrained Transformers}
We used pretrained transformer models for both the \ac{NER} and \ac{NEL} tasks. Given the highly specialized terminology and complex language of GutBrainIE documents, we focused on models pretrained on biomedical corpora, which are better suited to capture the linguistic and semantic characteristics of this collection.

We selected a total of seven transformer-based \acp{LM}:
\begin{itemize}
    \item \texttt{BioLinkBERT-base} and \texttt{BioLinkBERT-large}, which are pretrained on PubMed abstracts and incorporate citation-link information during pretraining~\cite{linkbert};
    
    \item \texttt{BiomedNLP-BiomedBERT-base-uncased-abstract} and \texttt{BiomedNLP-BiomedBERT-large-uncased-abstract}, which are BERT-based models trained from scratch on PubMed abstracts~\cite{pubmedbert}. We also included \texttt{BiomedNLP-BiomedBERT-large-uncased-abstract-fulltext}, a variant trained on both PubMed abstracts and full-text articles;
    
    \item \texttt{BiomedNLP-BiomedElectra-base-uncased-abstract} and \texttt{BiomedNLP-BiomedElectra-large-uncased-abstract}, which are ELECTRA-based models trained from scratch on PubMed abstracts~\cite{pubmedbert}.
\end{itemize}

The selection was guided by the results obtained by team \textit{GutInstincts} in the 2025 edition of the GutBrainIE challenge. Since these models achieved the best results in their experiments, we did not consider additional pretrained transformers~\cite{team_Gut-Instincts_paper}.

\subsection{Named Entity Recognition (NER) Module}
\label{sec:ner-module}

The first module of \ac{TWIX}, shown in Figure~\ref{fig:pipelineOverview}a, performs the identification and classification of entity mentions in the input documents. 
Instead of formulating \ac{NER} as a single-step classification problem, we adopt an \ac{EFCL} architecture, inspired by the approach proposed in~\cite{shlyk-etal-2026-mind}. While~\cite{shlyk-etal-2026-mind} implements \ac{EFCL} through prompt-based \acp{LLM}, our approach implements a similar principle using fine-tuned biomedical transformers. Specifically, our \ac{EFCL} module is a two-stage sequential architecture composed of a \ac{TE} module followed by a \ac{TC} module. The rationale is to separate mention detection from entity type assignment: the \ac{TE} module first identifies candidate spans that may correspond to relevant biomedical mentions, while the \ac{TC} module subsequently validates each candidate span and assigns it an entity type.

We frame \ac{TE} as a token classification task using the \ac{BIO} tagging schema~\cite{ramshaw-marcus-1995-text}. Differently from standard \ac{NER}, however, the label set does not include the 13 entity types defined by GutBrainIE. Instead, it contains a single generic \texttt{term} label. Each token is therefore assigned one of three possible tags: \texttt{B-term}, indicating the beginning of a candidate term; \texttt{I-term}, indicating that the token is inside a candidate term; or \texttt{O}, indicating that the token does not belong to any candidate term. 
The architecture consists of a pretrained biomedical transformer \ac{LM} followed by a token-level classification head. Given an input sequence, the transformer produces a contextualized representation for each token, which is then passed to a dense layer that independently predicts a probability distribution over the possible \ac{BIO} tags. Since the \ac{TE} module only distinguishes candidate terms from non-term tokens, the classification head has three output units, corresponding to \texttt{B-term}, \texttt{I-term}, and \texttt{O}.
After inference, consecutive tokens predicted as \texttt{B-term} or \texttt{I-term} are merged into candidate term spans.

The candidate spans produced by the \ac{TE} module are then passed to the \ac{TC} module, which we formulate as a sequence classification task. For each candidate term, the model receives as input the document text with the candidate span enclosed between two special markers, \texttt{[E1]} and \texttt{[/E1]}. The hidden representation corresponding to the \texttt{[E1]} marker is extracted and passed through a dense classification head. The classifier predicts either one of the GutBrainIE entity labels or a special \textit{not an entity} class. Candidate terms assigned to \textit{not an entity} are discarded, while the remaining ones are retained as final entity mentions with the label predicted by the classifier.

This two-stage design makes the \ac{NER} module more conservative than a standard single-step classifier. A false positive span can be filtered out at two different points: first, by not being extracted by the \ac{TE} module, and second, by being rejected by the \ac{TC} module through the \textit{not an entity} class. As a result, our \ac{EFCL} architecture is explicitly designed to improve precision, while still allowing the \ac{TE} component to operate with high recall over potentially relevant biomedical terms. Ideally, our architecture should preserve recall, since the entities filtered out by a generic entity/non-entity classifier (as used in the \ac{TE} component) should largely overlap with those removed by a more fine-grained classifier.

\subsection{Named Entity Linking (NEL) Module}
\label{sec:nel-module}

The entity mentions extracted and classified by the \ac{NER} module are passed to the \ac{NEL} module, whose goal is to assign each mention to a concept identifier from the biomedical resources defined in GutBrainIE. As shown in Figure~\ref{fig:pipelineOverview}b, we model \ac{NEL} as a two-stage pipeline composed of a candidate retriever followed by a reranker. Given an entity mention $e$ occurring in a textual context $m$, the retriever first selects a ranked set of the top-$k$ candidate concepts from the reference vocabularies. The reranker then scores these candidates in context and selects the most likely concept as the final prediction. 

We implemented this module using the \textit{entity-linkings} library, a unified framework for \ac{NEL} that provides multiple candidate retrievers and reranking models, allowing different combinations of methods to be tested within the same pipeline~\cite{sawada-etal-2026-entity}.

\paragraph{Candidate Retrieval.}
For the retrieval stage, we experimented with dual-encoder and text embedding models~\cite{wu-etal-2020-scalable,wang2024textembeddingsweaklysupervisedcontrastive}. Both approaches follow the same general principle: the mention and each candidate concept are mapped into a shared vector space, and candidates are ranked according to the similarity between their representations. The main difference between the two retrievers is in how these representations are computed.

Let $e$ be an entity mention occurring in context $m$, and let $c_i$ be a candidate concept. We represent each concept $c_i$ by a textual description $d(c_i)$ obtained by concatenating its preferred name and, when available, its definition or description from the corresponding knowledge base. The retriever computes a score $s_r(e,c_i)$ for each candidate concept and returns the candidates with the highest scores.

In the dual-encoder setting, the mention and the candidate concept are encoded separately, either by two different transformer encoders or by two branches of the same encoder. 
We use $f_{\theta}^{m}$ to indicate the mention encoder and $f_{\theta}^{c}$ to denote the candidate concept encoder.
The mention representation is obtained by inserting special markers around the mention in its context, similarly to the \ac{TC} module:
\[
m(e) = m_b \; \texttt{[E1]} \; e \; \texttt{[/E1]} \; m_a ,
\]
where $m_b$ and $m_a$ indicate the portions of context before and after the mention, respectively. The hidden representation of the \texttt{[E1]} marker is used as the mention vector:
\[
\mathbf{y}_e = f_{\theta}^{m}(m(e)).
\]
The candidate concept is encoded separately from its textual description:
\[
\mathbf{y}_{c_i} = f_{\theta}^{c}(d(c_i)).
\]
The retrieval score is then computed as the dot product between the mention and concept vectors:
\[
    s_r(e,c_i) = \mathbf{y}_e^\top \mathbf{y}_{c_i}.
\]

The model is trained with a contrastive objective that increases the score of the correct concept while decreasing the scores of negative candidates~\cite{khosla2020supervised}. Given a batch of $B$ mention-concept pairs $\{(e_j,c_j)\}_{j=1}^{B}$, the other concepts in the batch are used as in-batch negatives. The loss for a mention $e_j$ is computed as:
\[
    \mathcal{L}_{r}(e_j,c_j) =
    - \log
    \frac{
        \exp(s_r(e_j,c_j))
    }{
        \sum\limits_{k=1}^{B} \exp(s_r(e_j,c_k))
    }.
\]
In addition to random in-batch negatives, the \textit{entity-linkings} library also supports hard negatives, obtained by retrieving highly ranked but incorrect candidates for each training example. At inference time, retrieval is performed as a maximum inner-product search between the mention vector and the vectors of candidate concepts~\cite{wu-etal-2020-scalable}.

Text embedding retrievers follow the same scoring and ranking logic, but differ in how representations are obtained. Instead of using separate mention-side and concept-side encoders, both the mention context and the candidate concept description are encoded with a shared text embedding model $g_{\theta}$:
\[
\mathbf{y}_e = g_{\theta}\,(m(e)), 
\qquad
\mathbf{y}_{c_i} = g_{\theta}\,(d(c_i)).
\]
The score is again computed as:
\[
    s_r(e,c_i) = \mathbf{y}_e^\top \mathbf{y}_{c_i}.
\]
Thus, dual encoders and text embedding retrievers are operationally very similar at inference time, as both rank candidate concepts by vector similarity. Their difference mainly concerns how the embedding space is learned and how mention and concept representations are produced~\cite{wang2024textembeddingsweaklysupervisedcontrastive}.

\paragraph{Candidate Reranking.}
For the reranking stage, we used cross-encoder models~\cite{logeswaran-etal-2019-zero}. Differently from retrievers, which encode mentions and concepts independently, a cross encoder jointly encodes each mention-candidate pair. Given a mention $e$ in context $m$ and a candidate concept $c_i$, we build the input sequence:
\[
x(e,c_i) =
\texttt{[CLS]} \; m_b \; \texttt{[E1]} \; e \; \texttt{[/E1]} \; m_a
\; \texttt{[SEP]} \; d(c_i) \; \texttt{[SEP]} .
\]
This sequence is passed to a transformer encoder, and the final hidden representation of the \texttt{[CLS]} token is used as the contextual representation of the mention-candidate pair:
\[
\mathbf{h}_{e,c_i} = f_{\theta}^{ce}(x(e,c_i))_{\texttt{[CLS]}}.
\]
Each candidate is then assigned a reranking score:
\[
    s_{ce}(e,c_i) = \mathbf{w}^{\top}\mathbf{h}_{e,c_i} + b,
\]
where $\mathbf{w}$ and $b$ are learned parameters. Given the candidate set $C_k(e)$ returned by the retriever for mention $e$, the reranker defines a probability distribution over candidates:
\[
    p(c_i \mid e, C_k(e)) =
    \frac{
        \exp(s_{ce}(e,c_i))
    }{
        \sum\limits_{c_j \in C_k(e)} \exp(s_{ce}(e,c_j))
    }
\]
The model is trained with a softmax cross-entropy loss over the retrieved candidate set, maximizing the probability assigned to the correct concept~\cite{logeswaran-etal-2019-zero}. At inference time, the selected concept is:
\[
    \hat{c} =
    \arg \max\limits_{c_i \in C_k(e)} s_{ce}(e,c_i).
\]

For both retrieval and reranking, we experimented with two pretrained biomedical transformers: \texttt{BioLinkBERT-base} and \texttt{BiomedNLP-BiomedBERT-base-uncased-abstract}~\cite{linkbert,pubmedbert}. These models were selected as lightweight biomedical backbones, balancing domain specialization and computational cost.

\subsection{Relation Extraction (RE) Module}
The RE module shown in Figure~\ref{fig:pipelineOverview}c takes the entities predicted by the NER or NEL module and identified triples within the document. When considering entity annotations from the NER module (Figure~\ref{fig:pipelineOverview}a), the RE module solves the Mention-Level RE subtask. On the other hand, when considering entity concepts predicted by the NEL module (Figure~\ref{fig:pipelineOverview}b) the RE module performs Concept-Level RE.

We propose a two-stage training of the official baseline, i.e. ATLOP~\cite{atlop}. ATLOP is a transformer-based \ac{DocRE} model exploiting localized context pooling and an adaptive threshold for multi-label classification. Our two-stage training is inspired by DREEAM's architecture~\cite{ma_etal-2023}, a teacher-student model based on ATLOP. In DREEAM, the student model is trained in two steps. Firstly, a short pre-training phase initialize the weights of the networks exploiting distantly supervised data and weakly supervised labels predicted by the teacher model. Subsequently, the student model is finetuned on the manual dataset to learn precise predictions. Thus, our RE module features ATLOP firstly pre-trained on a large-scale (but noisy) set of documents for a small number of epochs and then fine tuned on a high-quality dataset. We employed \texttt{RoBERTa-large} as the transformer backbone of ATLOP.

\section{Experiments and Results}
\label{sec:experiments-and-results}

\subsection{Setup}
\label{sec:setup}
Training and inference jobs for \ac{NER} and \ac{NEL} are conducted on a MacStudio equipped with a M3 ultra chip and 512 GB of VRAM, while \ac{RE} jobs are done on a \ac{HPC} cluster, with jobs allocated access to one Nvidia A40 and 64 GB of RAM.

In terms of training data, the \ac{EFCL} \ac{NER} module uses both silver and gold folds, while the \ac{NEL} module uses only the gold fold. The \ac{RE} module leverages gold, silver, and bronze annotations.

For what concerns training hyper-parameters, \ac{TE} is trained using the AdamW optimizer for 30 epochs with a batch size of 16 and a learning rate of $2 \cdot 10^{-5}$~\cite{loshchilov2019decoupledweightdecayregularization}, and for each document titles and abstracts are concatenated to a single sequence
For what concerns the \ac{TC} modules, these are trained for 10 epochs with a batch size of 16 and a learning rate of $10^{-6}$. In the training of \ac{TC} modules, we also introduced synthetically generated negative samples. For each positive example of the training and dev sets we introduced a negative sample, making the model more robust in detecting false positives predicted by the upstream Term Extractor. These negative samples can have a length comprised from 1 to 5 words.

For \ac{NEL}, retrievers and rerankers were trained with different hyperparameter settings. Retriever models were trained for 5 epochs with a training batch size of 16, an evaluation batch size of 32, and no gradient accumulation. The number of soft negatives for each positive mention-concept pair was fixed to 10, while no hard negatives were used during retriever training.
Reranker models were trained for 4 epochs with a training batch size of 8, an evaluation batch size of 16, and gradient accumulation over 2 steps. For each mention, the reranker received up to 20 candidate concepts produced by the retrieval stage, with candidate retrieval performed using a batch size of 32. 

The RE module is trained the AdamW optimizer with a batch size of 4 and a learning rate of $5 \cdot 10^{-5}$. We tested different number of epochs both for the pre-training and fine-tuning phase and kept the best model based on the highest micro-F1 on the development dataset. We also tried two dataset splits, the first considers as distant data the silver and bronze-standard annotations (``\texttt{distant\textunderscore sb}") and the gold-standard annotations (``\texttt{manual\textunderscore g}") as high-quality dataset. The second considers as distant data the bronze-standard annotations (``\texttt{distant\textunderscore b}") and the silver and gold-standard annotations as high-quality dataset (``\texttt{manual\textunderscore gs}").


\subsection{Experiments on the Development Dataset}
This Section presents the experimental results obtained on the development set across all subtasks. 
These results motivated the selection of the final runs submitted to the GutBrainIE challenge for evaluation on the test set. 
All results are computed using the official evaluation script provided by the challenge organizers. 
We report component-level analyses aimed at isolating the contribution and limitations of individual modules. In particular, we evaluate the \ac{TE} module independently by measuring span detection performance, and we assess the \ac{TC}, \ac{NEL}, and \ac{RE} modules provided with ground-truth entity mentions instead of predictions from upstream components. These analyses allow us to better understand how errors propagate across \ac{TWIX} and to identify the main bottlenecks affecting downstream performance.
Throughout the analysis, we focus on micro-averaged metrics, which better account for the strong class imbalance characterizing both entity labels and relation predicates.

\begin{table}[t]
\caption{Micro-averaged \ac{NER} performance on the development set for the seven selected pretrained biomedical transformers, evaluated under the standard token-classification formulation. Performance is reported in terms of Precision ($P$), Recall ($R$), and $F_1$ score, and runs are sorted by decreasing $F_1$. Best results for each metric are bolded and second-best results are underlined.}
\label{tab:NER-single-dev-scores}
\centering
\begin{tabular}{p{0.62\textwidth}ccc}
\toprule
Model & $P$ & $R$ & $F_1$ \\
\midrule
BiomedNLP-BiomedBERT-base-uncased-abstract-fulltext & 0.8009 & 0.8235 & \textbf{0.8120} \\
BioLinkBERT-base & 0.7972 & \underline{0.8267} & \underline{0.8117} \\
BiomedNLP-BiomedBERT-large-uncased-abstract & 0.7877 & \textbf{0.8314} & 0.8090 \\
BiomedNLP-BiomedElectra-large-uncased-abstract & \underline{0.8070} & 0.8096 & 0.8083 \\
BiomedNLP-BiomedBERT-base-uncased-abstract & \textbf{0.8159} & 0.8001 & 0.8079 \\
BioLinkBERT-large & 0.7783 & 0.8259 & 0.8014 \\
BiomedNLP-BiomedElectra-base-uncased-abstract & 0.7791 & 0.8171 & 0.7977 \\
\bottomrule
\end{tabular}
\end{table}

\subsubsection{Named Entity Recognition}
\label{sec:ner-dev-results}

We evaluated three families of approaches for the \ac{NER} subtask on the development set: (i) standard transformer-based \ac{NER} models framed as token classification, (ii) post-processing ensembles of these \ac{NER} models, and (iii) our proposed two-stage \ac{EFCL} architecture.
The models in (i) were trained with the same hyper-parameters used for the \ac{TE} modules (see Section~\ref{sec:setup}).
This comparison is motivated by the results of the previous GutBrainIE edition, where transformer-based \ac{NER} systems and ensemble strategies proved to be among the most effective approaches for entity recognition, where biomedical transformer models and ensemble strategies proved to be among the most effective solutions for entity recognition~\cite{martinelli2025overview,team_GutUZH_paper,team_Gut-Instincts_paper}. Therefore, we use these models as reference systems against which to assess the benefits of our \ac{EFCL} design.

The standard \ac{NER} baselines follow the classical token-classification formulation with the \ac{BIO} tagging schema~\cite{ramshaw-marcus-1995-text}. Each model consists of a pretrained biomedical transformer followed by a dense classification head predicting \texttt{B-<label>}, \texttt{I-<label>}, or \texttt{O} tags for each token. 
Consecutive tokens are merged into an entity mention when a \texttt{B-<label>} tag is followed by consecutive \texttt{I-<label>} tags associated with the same entity label.
When an \texttt{I-<label>} tag follows a tag associated with a different entity type, it is treated as the beginning of a new entity mention. 

Table~\ref{tab:NER-single-dev-scores} reports the performance of the seven selected biomedical transformers. Overall, the results show that no model clearly outperforms the others. 
The best model, \texttt{BiomedNLP-BiomedBERT-base-uncased-abstract-fulltext}, reaches a micro-averaged $F_1$ of 0.8120, while the lowest-scoring model, \texttt{BiomedNLP-BiomedElectra-base-uncased-abstract}, obtains 0.7977. 
The difference between the best and worst model is therefore only 0.0143 $F_1$ points, indicating that, under this standard token-classification setting, model choice alone has a limited impact. A similar pattern can be observed for precision and recall: some models are slightly more recall-oriented, such as \texttt{BiomedNLP-BiomedBERT-large-uncased-abstract}, while others obtain slightly higher precision, such as \texttt{BiomedNLP-BiomedBERT-base-uncased-abstract}. However, these differences do not translate into a clearly superior model.

We then evaluated post-processing majority-voting ensembles of these standard \ac{NER} models~\cite{team_Gut-Instincts_paper}. In this setting, predictions produced by different models are combined after inference, without modifying the underlying architectures. We considered ensembles of different sizes, ranging from two to seven models. A predicted entity mention is retained when it is predicted by the majority of the models included in the ensemble, requiring an exact match in terms of start offset, end offset, and entity label. 
Table~\ref{tab:ner_majority_voting_ensembles} reports the top eight majority-voting ensembles under micro-averaged $F_1$ score.

\begin{table*}[t]
\centering
\scriptsize
\setlength{\tabcolsep}{3pt}
\renewcommand{\arraystretch}{1.25}
\caption{Micro-averaged \ac{NER} performance of the top eight majority-voting ensembles on the development set. The \textit{Size} column reports the number of models included in the ensemble, while the \textit{Models} column lists the corresponding pretrained transformer backbones. Performance is reported in terms of precision ($P$), recall ($R$), and $F_1$ score, with runs sorted by decreasing $F_1$. Best results for each metric are bolded and second-best results are underlined.}
\label{tab:ner_majority_voting_ensembles}
\begin{tabular}{ccp{0.50\textwidth}ccc}
\toprule
ID & Size & Models & $P$ & $R$ & $F_1$ \\
\midrule

1 & 4 &
\begin{tabular}[t]{@{}l@{}}
BioLinkBERT-base \\
BiomedNLP-BiomedBERT-base-uncased-abstract-fulltext \\
BiomedNLP-BiomedBERT-base-uncased-abstract \\
BiomedNLP-BiomedBERT-large-uncased-abstract
\end{tabular}
& 0.8835 & 0.8120 & \textbf{0.8462} \\
\midrule

2 & 5 &
\begin{tabular}[t]{@{}l@{}}
BioLinkBERT-large \\
BiomedNLP-BiomedBERT-base-uncased-abstract \\
BiomedNLP-BiomedBERT-large-uncased-abstract \\
BiomedNLP-BiomedElectra-base-uncased-abstract \\
BiomedNLP-BiomedElectra-large-uncased-abstract
\end{tabular}
& 0.8623 & 0.8294 & \underline{0.8455} \\
\midrule

3 & 6 &
\begin{tabular}[t]{@{}l@{}}
BioLinkBERT-base \\
BiomedNLP-BiomedBERT-base-uncased-abstract-fulltext \\
BiomedNLP-BiomedBERT-base-uncased-abstract \\
BiomedNLP-BiomedBERT-large-uncased-abstract \\
BiomedNLP-BiomedElectra-base-uncased-abstract \\
BiomedNLP-BiomedElectra-large-uncased-abstract
\end{tabular}
& 0.8761 & 0.8163 & 0.8452 \\
\midrule

4 & 5 &
\begin{tabular}[t]{@{}l@{}}
BioLinkBERT-base \\
BioLinkBERT-large \\
BiomedNLP-BiomedBERT-base-uncased-abstract-fulltext \\
BiomedNLP-BiomedBERT-base-uncased-abstract \\
BiomedNLP-BiomedBERT-large-uncased-abstract
\end{tabular}
& 0.8591 & \underline{0.8318} & 0.8452 \\
\midrule

5 & 5 &
\begin{tabular}[t]{@{}l@{}}
BioLinkBERT-base \\
BioLinkBERT-large \\
BiomedNLP-BiomedBERT-base-uncased-abstract-fulltext \\
BiomedNLP-BiomedBERT-large-uncased-abstract \\
BiomedNLP-BiomedElectra-large-uncased-abstract
\end{tabular}
& 0.8569 & \textbf{0.8338} & 0.8452 \\
\midrule

6 & 4 &
\begin{tabular}[t]{@{}l@{}}
BioLinkBERT-base \\
BiomedNLP-BiomedBERT-base-uncased-abstract-fulltext \\
BiomedNLP-BiomedBERT-base-uncased-abstract \\
BiomedNLP-BiomedElectra-large-uncased-abstract
\end{tabular}
& \textbf{0.8946} & 0.8009 & 0.8451 \\
\midrule

7 & 4 &
\begin{tabular}[t]{@{}l@{}}
BioLinkBERT-base \\
BiomedNLP-BiomedBERT-base-uncased-abstract \\
BiomedNLP-BiomedBERT-large-uncased-abstract \\
BiomedNLP-BiomedElectra-large-uncased-abstract
\end{tabular}
& \underline{0.8909} & 0.8036 & 0.8450 \\
\midrule

8 & 4 &
\begin{tabular}[t]{@{}l@{}}
BioLinkBERT-base \\
BiomedNLP-BiomedBERT-base-uncased-abstract \\
BiomedNLP-BiomedBERT-large-uncased-abstract \\
BiomedNLP-BiomedElectra-base-uncased-abstract
\end{tabular}
& 0.8827 & 0.8092 & 0.8444 \\
\bottomrule

\end{tabular}
\end{table*}

Consistently with the results observed in the previous edition of GutBrainIE, ensembling substantially improves over individual transformer models~\cite{martinelli2025overview}. The best ensemble obtains 0.8462 $F_1$, corresponding to an absolute improvement of 0.0342 over the best single-model baseline. This improvement is mainly given by a large increase in Precision, which rises from 0.8009 for the best single model to 0.8835 for the best ensemble, while recall remains comparable. This confirms that majority voting is effective in filtering out predictions that are not consistently supported across models. 
At the same time, the analysis of the best ensemble configurations shows that the improvement is not driven by a single clearly superior backbone. 
Among the configurations reaching the highest $F_1$ values, the most recurrent models are \texttt{BioLinkBERT-base}, \texttt{BiomedNLP-BiomedBERT-base-uncased-abstract}, and \texttt{BiomedNLP-BiomedBERT-large-uncased-abstract}, but several other models also appear in competitive combinations. Moreover, the best results are obtained mostly with ensembles of size four or five, indicating that adding more models does not necessarily lead to better performance~\cite{team_Gut-Instincts_paper}.

Finally, Table~\ref{tab:termMerged-dev-top15} reports the top 15 configurations (under micro-$F_1$) obtained with our proposed \ac{EFCL} architecture, described in Section~\ref{sec:ner-module}.
The full results are reported in Table~\ref{tab:termMerged-dev-full} in the Appendix.
Our two-stage \ac{TE}+\ac{TC} design consistently outperforms both standard \ac{NER} models and majority-voting ensembles. The best configurations, obtained with \texttt{BiomedNLP-BiomedBERT-large-uncased-abstract} as term extractor and either \texttt{BiomedNLP-BiomedBERT-large-uncased-abstract} or \texttt{BiomedNLP-BiomedBERT-base-uncased-abstract-fulltext} as term classifier, reach a micro-averaged $F_1$ of 0.8925. This corresponds to an absolute improvement of 0.0805 over the best single transformer-based \ac{NER} model and 0.0463 over the best majority-voting ensemble.

\begin{table}[t]
\caption{Micro-averaged \ac{NER} performance on the development set for the proposed \ac{EFCL} architecture. Each row reports a combination of \ac{TE} and \ac{TC} models, with runs sorted by decreasing $F_1$ score. Performance is reported in terms of precision ($P$), recall ($R$), and $F_1$ score. Best results for each metric are bolded and second-best results are underlined.}
\label{tab:termMerged-dev-top15}
\resizebox{\textwidth}{!}{%
\begin{tabular}{lllccc}
\toprule
ID & Term Extractor Model                                & Term Classifier Model                               & $P$    & $R$    & $F_1$  \\ \midrule
1  & BiomedNLP-BiomedBERT-large-uncased-abstract         & BiomedNLP-BiomedBERT-large-uncased-abstract         & 0.9544 & \textbf{0.8382} & \textbf{0.8925} \\ \midrule
2  & BiomedNLP-BiomedBERT-large-uncased-abstract         & BiomedNLP-BiomedBERT-base-uncased-abstract-fulltext & 0.9544 & \textbf{0.8382} & \textbf{0.8925} \\ \midrule
3  & BiomedNLP-BiomedBERT-large-uncased-abstract         & BiomedNLP-BiomedBERT-base-uncased-abstract          & 0.9507 & \underline{0.8346} & \underline{0.8889} \\ \midrule
4  & BiomedNLP-BiomedBERT-large-uncased-abstract         & BioLinkBERT-large                                   & 0.9494 & 0.8338 & 0.8879 \\ \midrule
5  & BiomedNLP-BiomedBERT-large-uncased-abstract         & BioLinkBERT-base                                    & 0.9485 & 0.8326 & 0.8868 \\ \midrule
6  & BiomedNLP-BiomedBERT-large-uncased-abstract         & BiomedNLP-BiomedElectra-base-uncased-abstract       & 0.9476 & 0.8326 & 0.8864 \\ \midrule
7  & BioLinkBERT-base                                    & BiomedNLP-BiomedBERT-large-uncased-abstract         & 0.9472 & 0.8322 & 0.8860 \\ \midrule
8  & BiomedNLP-BiomedElectra-large-uncased-abstract      & BiomedNLP-BiomedBERT-large-uncased-abstract         & 0.9516 & 0.8259 & 0.8843 \\ \midrule
9  & BiomedNLP-BiomedBERT-large-uncased-abstract         & BiomedNLP-BiomedElectra-large-uncased-abstract      & \textbf{0.9574} & 0.8207 & 0.8838 \\ \midrule
10 & BioLinkBERT-base                                    & BiomedNLP-BiomedBERT-base-uncased-abstract-fulltext & 0.9511 & 0.8255 & 0.8838 \\ \midrule
11 & BiomedNLP-BiomedBERT-base-uncased-abstract-fulltext & BiomedNLP-BiomedBERT-large-uncased-abstract         & 0.9535 & 0.8219 & 0.8828 \\ \midrule
12 & BiomedNLP-BiomedElectra-large-uncased-abstract      & BiomedNLP-BiomedBERT-base-uncased-abstract-fulltext & 0.9561 & 0.8199 & 0.8828 \\ \midrule
13 & BiomedNLP-BiomedBERT-base-uncased-abstract-fulltext & BiomedNLP-BiomedBERT-base-uncased-abstract-fulltext & 0.9531 & 0.8215 & 0.8824 \\ \midrule
14 & BioLinkBERT-large                                   & BiomedNLP-BiomedBERT-base-uncased-abstract-fulltext & \underline{0.9572} & 0.8159 & 0.8809 \\ \midrule
15 & BioLinkBERT-base                                    & BiomedNLP-BiomedElectra-large-uncased-abstract      & 0.9479 & 0.8227 & 0.8809 \\
\bottomrule
\end{tabular}%
}
\end{table}

The main gains of the \ac{EFCL} architecture come from precision. The best \ac{TE}+\ac{TC} configuration obtains 0.9544 precision, compared with 0.8009 for the best single model and 0.8835 for the best ensemble. This behavior is consistent with the design of the proposed architecture. In the standard \ac{NER} setting, each token is directly assigned an entity label or the \texttt{O} label by a single model. In contrast, \ac{EFCL} introduces two sequential filtering decisions. First, the \ac{TE} module decides whether a span should be extracted as a candidate term. Then, the \ac{TC} module assigns an entity type to the candidate term or discards it by predicting the \textit{not an entity} class. As a consequence, false positive spans can be removed at two different stages of the pipeline, which explains the strong increase in precision. Importantly, this precision gain does not come at the cost of lower recall: the best \ac{TE}+\ac{TC} system obtains a recall of 0.8382, which is also higher than both the best single model and the best ensemble.

To better understand the contribution of the two components, we also evaluated the \ac{TE} and \ac{TC} modules separately. For \ac{TE}, we compared predicted spans against the ground-truth entities of the development set while ignoring entity labels, thus measuring the ability of the extractor to identify relevant mention boundaries independently of the downstream classification step. Results are reported in Table~\ref{tab:termExtractor-dev}. As observed for standard \ac{NER}, all seven biomedical transformers obtain competitive results, with no model clearly dominating the others. The best configuration, based on \texttt{BiomedNLP-BiomedBERT-large-uncased-abstract}, reaches 0.8549 $F_1$, but the remaining models are close, with the lowest score still equal to 0.8440. This confirms that the selected transformers behave similarly also when the task is reduced to label-agnostic \acf{TE}.

\begin{table}[t]
\caption{Micro-averaged span-detection performance of the \ac{TE} module on the development set. Runs are sorted by decreasing $F_1$ score. Performance is reported in terms of precision ($P$), recall ($R$), and $F_1$ score. Best results for each metric are bolded and second-best results are underlined.}
\label{tab:termExtractor-dev}
\resizebox{0.7\textwidth}{!}{%
\begin{tabular}{lccc}
\toprule
Term Extractor Model & $P$ & $R$ & $F_1$ \\ 
\midrule
BiomedNLP-BiomedBERT-large-uncased-abstract         & 0.8324 & \textbf{0.8786} & \textbf{0.8549} \\ \midrule
BiomedNLP-BiomedBERT-base-uncased-abstract-fulltext & 0.8420 & 0.8624 & \underline{0.8520} \\ \midrule
BioLinkBERT-large                                   & 0.8501 & 0.8528 & 0.8515 \\ \midrule
BiomedNLP-BiomedElectra-large-uncased-abstract      & 0.8429 & 0.8580 & 0.8504 \\ \midrule
BiomedNLP-BiomedElectra-base-uncased-abstract       & \underline{0.8505} & 0.8485 & 0.8495 \\ \midrule
BioLinkBERT-base                                    & 0.8254 & \underline{0.8683} & 0.8463 \\ \midrule
BiomedNLP-BiomedBERT-base-uncased-abstract          & \textbf{0.8600} & 0.8286 & 0.8440 \\ 
\bottomrule
\end{tabular}%
}
\end{table}

We then evaluated the \ac{TC} module in isolation by providing it with ground-truth entity mentions from the development set. After inference, mentions classified as \textit{not an entity} were removed, and the remaining predictions were evaluated against the gold labels. Results are reported in Table~\ref{tab:termClassifier-dev}. Also in this setting the seven models obtain very similar scores, with all configurations reaching an $F_1$ higher than 0.93. Nevertheless, consistently with the results obtained with the full \ac{EFCL} pipeline, the best-performing models belong to the \texttt{BiomedBERT} family. In particular, \texttt{BiomedNLP-BiomedBERT-large-uncased-abstract} and \texttt{BiomedNLP-BiomedBERT-base-uncased-abstract-fulltext} obtain the two highest $F_1$ scores for both the isolated \ac{TC} module and the full merged architecture. This suggests that, although all selected biomedical transformers are competitive, \texttt{BiomedBERT}-based models provide the most stable performance in both stages of the proposed architecture.

\begin{table}[t]
\caption{Micro-averaged entity classification performance of the \ac{TC} module on the development set. The classifier is provided with ground-truth entity mentions and evaluated on its ability to assign the correct GutBrainIE entity label. Runs are sorted by decreasing $F_1$ score. Performance is reported in terms of precision ($P$), recall ($R$), and $F_1$ score. Best results for each metric are bolded and second-best results are underlined.}
\label{tab:termClassifier-dev}
\resizebox{0.7\textwidth}{!}{%
\begin{tabular}{lccc}
\toprule
Term Classifier Model & $P$ & $R$ & $F_1$ \\ 
\midrule
BiomedNLP-BiomedBERT-large-uncased-abstract         & \underline{0.9485} & \textbf{0.9349} & \textbf{0.9417} \\ \midrule
BiomedNLP-BiomedBERT-base-uncased-abstract-fulltext & \textbf{0.9488} & \underline{0.9338} & \underline{0.9412} \\ \midrule
BiomedNLP-BiomedBERT-base-uncased-abstract          & 0.9451 & 0.9290 & 0.9370 \\ \midrule
BiomedNLP-BiomedElectra-large-uncased-abstract      & 0.9436 & 0.9290 & 0.9362 \\ \midrule
BioLinkBERT-base                                    & 0.9409 & 0.9290 & 0.9349 \\ \midrule
BiomedNLP-BiomedElectra-base-uncased-abstract       & 0.9404 & 0.9270 & 0.9337 \\ \midrule
BioLinkBERT-large                                   & 0.9428 & 0.9226 & 0.9326 \\ 
\bottomrule
\end{tabular}%
}
\end{table}

Overall, these results support the choice of the \ac{EFCL} architecture as our final \ac{NER} module. Although standard biomedical transformers provide strong and stable baselines, their performance remains clustered in a narrow range. Ensembling improves effectiveness, especially by increasing Precision, but remains clearly below our two-stage approach. The proposed \ac{EFCL} design better matches the needs of the GutBrainIE \ac{NER} task, where avoiding incorrect entity predictions is crucial because downstream \ac{NEL} and \ac{RE} modules directly depend on the quality of the extracted mentions.

\subsubsection{Named Entity Recognition and Disambiguation (Subtask 6.1.2)}
\label{sec:nerd-dev-results}

The \ac{NERD} subtask extends \ac{NER} by requiring each predicted entity mention to be linked to a concept identifier. In \ac{TWIX}, this task is addressed by applying the \ac{NEL} module to the entity mentions produced by an upstream \ac{NER} system. The \ac{NEL} module always assigns exactly one concept to each entity mention it receives. Therefore, it does not perform any additional filtering or correction of the upstream \ac{NER} predictions. As a consequence, the final \ac{NERD} performance depends both on the quality of the extracted entity mentions and on the ability of the \ac{NEL} module to assign the correct concept.

To isolate the performance of the \ac{NEL} component from possible bottlenecks introduced by \ac{NER}, we evaluated the \ac{NEL} configurations on the development set using ground-truth entity mentions, similarly to what we have done to assess the effectiveness of the \ac{TC} module in isolation. 
This setting allows us to estimate the linking performance under gold entity boundaries and labels, without errors propagated from the upstream \ac{NER} module. Since the input entities exactly match the ground truth and the \ac{NEL} module assigns one concept to each of them, Precision, Recall, and $F_1$ coincide: each wrong concept assignment produces one false positive and one false negative, while each correct assignment produces one true positive.

As described in Section~\ref{sec:methodology}, our \ac{NEL} module is composed of two sequential components: a retriever and a reranker. The retriever produces a ranked list of the top-$k$ candidate concepts for each entity mention, with $k=10$ in our experiments. The reranker then selects the most probable concept among these candidates. We experimented with two retriever families, namely dual encoders and text embedding models, as detailed in Section~\ref{sec:nel-module}. 
For the reranking stage, we employed cross-encoder models.

Given the results obtained for \ac{NER}, where no pretrained transformer clearly outperformed the others, we restricted \ac{NEL} experiments to two base biomedical transformers: \texttt{BiomedNLP-BiomedBERT-base-uncased-abstract}, reported as \texttt{BiomedBERT}, and \texttt{BioLinkBERT-base}, reported as \texttt{BioLinkBERT}. We selected the base versions to limit training and inference costs, since \ac{NEL} is computationally more demanding than \ac{NER}: candidate retrieval requires encoding mentions and concepts, while reranking requires evaluating mention-candidate pairs with a cross encoder.

Table~\ref{tab:nerd-dev} reports the development-set results obtained by all tested retriever-reranker configurations. Overall, the differences among configurations are small. The best result is obtained by using a text embedding retriever based on \texttt{BiomedBERT} and a cross-encoder reranker also based on \texttt{BiomedBERT}, reaching a micro-averaged $F_1$ of 0.8036. However, several other configurations obtain very similar results, with most scores clustered around 0.80. The difference between the best and worst configuration is only 0.0119 $F_1$ points, suggesting that neither the retriever family nor the specific transformer backbone has a strong impact in this setting.

\begin{table}[t]
\caption{Micro-averaged \ac{NERD} performance on the development set obtained by providing the \ac{NEL} module with ground-truth entity mentions. This setting isolates concept-linking performance from upstream \ac{NER} errors. Since each input mention is linked to exactly one concept, and no mentions are added or removed, Precision ($P$), Recall ($R$), and $F_1$ score coincide. Runs are sorted by decreasing $F_1$. Best results for each metric are bolded and second-best results are underlined.}
\label{tab:nerd-dev}
\resizebox{0.8\textwidth}{!}{%
\begin{tabular}{llllccc}
\toprule
Retriever      & Retriever Model & Reranker      & Reranker Model & $P$    & $R$    & $F_1$  \\ 
\midrule
\multirow{4}{*}{Text Embedding} 
& \multirow{2}{*}{BiomedBERT} 
& Cross Encoder & BiomedBERT  
& \textbf{0.8036} & \textbf{0.8036} & \textbf{0.8036} \\

& & Cross Encoder & BioLinkBERT 
& \underline{0.8017} & \underline{0.8017} & \underline{0.8017} \\

\cmidrule{2-7}

& \multirow{2}{*}{BioLinkBERT} 
& Cross Encoder & BiomedBERT  
& 0.7933 & 0.7933 & 0.7933 \\

& & Cross Encoder & BioLinkBERT 
& 0.7917 & 0.7917 & 0.7917 \\ 

\midrule

\multirow{4}{*}{Dual Encoder}   
& \multirow{2}{*}{BiomedBERT} 
& Cross Encoder & BiomedBERT  
& \underline{0.8017} & \underline{0.8017} & \underline{0.8017} \\ 

& & Cross Encoder & BioLinkBERT 
& 0.8013 & 0.8013 & 0.8013 \\

\cmidrule{2-7}

& \multirow{2}{*}{BioLinkBERT} 
& Cross Encoder & BiomedBERT  
& \underline{0.8017} & \underline{0.8017} & \underline{0.8017} \\

& & Cross Encoder & BioLinkBERT 
& 0.7941 & 0.7941 & 0.7941 \\

\bottomrule
\end{tabular}%
}
\end{table}

Based on these results, we selected a compact set of \ac{NEL} configurations for the final test submissions. Since all configurations performed similarly, we fixed the retriever family to text embedding and the reranker family to cross encoder, while varying the transformer backbone used by each component. This produced four \ac{NEL} configurations: \texttt{BiomedBERT}-\texttt{BiomedBERT}, \texttt{BiomedBERT}-\texttt{BioLinkBERT}, \texttt{BioLinkBERT}-\texttt{BiomedBERT}, and \texttt{BioLinkBERT}-\texttt{BioLinkBERT}. We then combined these four linking configurations with the six best \ac{NER} configurations obtained with our two-stage \ac{EFCL} architecture, resulting in 24 submitted runs for the \ac{NERD} subtask.

\subsubsection{Relation Extraction}
\label{sec:re-dev-results}
We evaluate ATLOP on the development dataset in the mention-level and concept-level RE task. We test different datasets different number of epochs for pre-training and fine-tuning. To isolate the performance of the \ac{RE} module from the possible bottleneck derived from the \ac{NER} and \ac{NERD} modules, we leverage the entity annotations in the development ground truth. Tables report the micro-averaged results.

Table~\ref{tab:mre-dev} reports the results on the development dataset for task 6.2.1 (mention-level RE). Changing the training dataset and the number of epochs influences the model's precision, but has not effect on recall. Given the same number of epochs for pre-training and fine-tuning, exploiting the gold annotations as manual dataset (``\texttt{manual\_g}") and the silver and bronze as distant (``\texttt{manual\_sb}") yields superior precision than using gold and silver annotations as manual dataset (``\texttt{manual\_gs}") and the bronze as distant (``\texttt{manual\_b}"). This shows that silver and bronze annotations are useful for pre-training, but gold annotations have a stronger impact on the quality of the predictions. Given the same training dataset, the best performing configuration is the one with the highest number of training epochs. Thus, we can conclude that longer training is beneficial to the model's precision. The results on the development dataset for task 6.2.2 (concept-level RE) reported in Table~\ref{tab:cre-dev} confirm the same trend.

\begin{table}[t]
\caption{Micro-averaged mention-Level RE performance on the development dataset. We report the training dataset (``Training Data"), the number of training epochs for the pre-training (``Pre-Training") and fine-tuning phase (``Fine-Tuning"). Best results are bolded and the second-best results are underlined.}
\label{tab:mre-dev}
\begin{tabular}{lccccc}
\toprule
Training Data & Pre-Training & Fine-Tuning & $P$ & $R$ & $F_1$ \\
\midrule
\multirow{3}{*}{{\begin{tabular}[c]{@{}l@{}}distant\_sb \\ manual\_g\end{tabular}}} & 2 & 10 & 0.7865 & \textbf{0.4205} & 0.5481 \\
& 5 & 50 & \underline{0.7878} & \textbf{0.4205} & \underline{0.5484} \\
& 10 & 100 & \textbf{0.8870} & \textbf{0.4205} & \textbf{0.5706} \\
\midrule
\multirow{3}{*}{{\begin{tabular}[c]{@{}l@{}}distant\_b \\ manual\_gs\end{tabular}}} & 2 & 10 & 0.7789 & \textbf{0.4205} & 0.5462 \\
& 5 & 50 & 0.7291 & \textbf{0.4205} & 0.5334 \\
& 10 & 100 & 0.7664 & \textbf{0.4205} & 0.5431 \\       
\bottomrule
\end{tabular}
\end{table}


\begin{table}[t]
\caption{Micro-averaged concept-Level RE performance on the development dataset. We report the training dataset (``Training Data"), the number of training epochs for the pre-training (``Pre-Training") and fine-tuning phase (``Fine-Tuning"). Best results are bolded and the second-best results are underlined.}
\label{tab:cre-dev}
\begin{tabular}{lccccc}
\toprule
Training Data & Pre-Training & Fine-tuning & $P$ & $R$ & $F_1$ \\
\midrule
\multirow{3}{*}{{\begin{tabular}[c]{@{}l@{}}distant\_sb \\ manual\_g\end{tabular}}} & 2 & 10 & 0.7912 & \textbf{0.4263} & 0.5541 \\
& 5 & 50 & \underline{0.7945} & \textbf{0.4263} & \underline{0.5549} \\
& 10 & 100 & \textbf{0.8773} & \textbf{0.4263} & \textbf{0.5738} \\
\midrule
\multirow{3}{*}{{\begin{tabular}[c]{@{}l@{}}distant\_b \\ manual\_gs\end{tabular}}} & 2 & 10 & 0.7762 & \textbf{0.4263} & 0.5505 \\
& 5 & 50 & 0.7359 & \textbf{0.4263} & 0.5399 \\
& 10 & 100 & 0.7657 & \textbf{0.4263} & 0.5477 \\       
\bottomrule
\end{tabular}
\end{table}

\section{Submission Results}
\label{sec:submission-results}

\begin{table}[t]
\caption{Micro-averaged \ac{NER} results on the official test set. Runs with IDs starting with \texttt{merged} correspond to our two-stage \ac{EFCL} architecture, where the system description reports the \ac{TE} model on the left and the \ac{TC} model on the right. Runs with IDs starting with \texttt{ensemble} correspond to majority-voting ensembles selected from the best development-set configurations, while runs with IDs starting with \texttt{ner} correspond to standard single-model transformer-based \ac{NER} systems. The official baseline is highlighted.}
\label{tab:ner-test}
\resizebox{\textwidth}{!}{%
\begin{tabular}{llccc}
\toprule
Run ID    & System Description                                                            & $P$    & $R$    & $F_1$  \\ \midrule
merged6   & BiomedBERT-large-uncased-abstract-W-BiomedBERT-large-uncased-abstract         & 0.9548 & 0.8058 & 0.8740 \\ \midrule
merged5   & BiomedBERT-large-uncased-abstract-W-BiomedBERT-base-uncased-abstract-fulltext & 0.9543 & 0.8058 & 0.8738 \\ \midrule
merged1   & BioLinkBERT-base-W-BiomedBERT-large-uncased-abstract                          & 0.9546 & 0.8021 & 0.8717 \\ \midrule
merged3   & BiomedBERT-large-uncased-abstract-W-BioLinkBERT-large                         & 0.9508 & 0.8021 & 0.8701 \\ \midrule
merged4   & BiomedBERT-large-uncased-abstract-W-BiomedBERT-base-uncased-abstract          & 0.9478 & 0.8002 & 0.8678 \\ \midrule
merged2   & BiomedBERT-large-uncased-abstract-W-BioLinkBERT-base                          & 0.9456 & 0.7984 & 0.8658 \\ \midrule
merged8   & BiomedElectra-large-uncased-abstract-W-BiomedBERT-large-uncased-abstract      & 0.9618 & 0.7835 & 0.8636 \\ \midrule
merged7   & BiomedBERT-large-uncased-abstract-W-BiomedElectra-base-uncased-abstract       & 0.9372 & 0.7910 & 0.8579 \\ \midrule
ensemble5 & majority-k5-combo0383                                                         & 0.8639 & 0.7884 & 0.8244 \\ \midrule
ensemble7 & majority-k4-combo0216                                                         & 0.8600 & 0.7898 & 0.8234 \\ \midrule
ensemble8 & majority-k4-combo0215                                                         & 0.8587 & 0.7906 & 0.8232 \\ \midrule
ensemble6 & majority-k4-combo0196                                                         & 0.8807 & 0.7713 & 0.8224 \\ \midrule
ensemble4 & majority-k5-combo0376                                                         & 0.8416 & 0.8036 & 0.8221 \\ \midrule
ensemble1 & majority-k4-combo0194                                                         & 0.8657 & 0.7809 & 0.8211 \\ \midrule
ensemble3 & majority-k6-combo0698                                                         & 0.8599 & 0.7850 & 0.8208 \\ \midrule
ensemble2 & majority-k5-combo0537                                                         & 0.8603 & 0.7832 & 0.8199 \\ \midrule
ner1      & BioLinkBERT-base                                                              & 0.7891 & 0.8128 & 0.8008 \\ \midrule
\rowcolor[HTML]{C9DAF8} 
BASELINE  & GLiNER                                                                        & 0.7782 & 0.8221 & 0.7996 \\ \midrule
ner3      & BiomedBERT-base-uncased-abstract                                              & 0.8046 & 0.7891 & 0.7968 \\ \midrule
ner4      & BiomedBERT-base-uncased-abstract-fulltext                                     & 0.7807 & 0.7917 & 0.7862 \\ \midrule
ner5      & BiomedBERT-large-uncased-abstract                                             & 0.7710 & 0.7976 & 0.7841 \\ \midrule
ner6      & BiomedElectra-base-uncased-abstract                                           & 0.7665 & 0.7984 & 0.7821 \\ \midrule
ner2      & BioLinkBERT-large                                                             & 0.7682 & 0.7958 & 0.7817 \\ \midrule
ner7      & BiomedElectra-large-uncased-abstract                                          & 0.7723 & 0.7895 & 0.7808 \\ \bottomrule
\end{tabular}%
}
\end{table}

\subsection{Named Entity Recognition (Subtask 6.1.1)}
\label{sec:ner-test-results}

For the \ac{NER} subtask, we submitted three groups of runs: the seven single-model transformer-based \ac{NER} systems, the top eight majority-voting ensembles selected on the development set, and the top eight configurations of our \ac{EFCL} architecture. Table~\ref{tab:ner-test} reports the micro-averaged results obtained on the official test set, together with the official baseline provided by the organizers.

The results confirm the trends already observed on the development set. Single-model transformer-based \ac{NER} systems obtain performances close to the baseline, with only \texttt{BioLinkBERT-base} marginally outperforming it. Specifically, \texttt{BioLinkBERT-base} reaches a micro-averaged $F_1$ of 0.8008, compared with 0.7996 obtained by the baseline. All the other single-model runs remain slightly below the baseline, with $F_1$ scores ranging between 0.7808 and 0.7968. This confirms that, in this setting, simply replacing the baseline with a biomedical transformer fine-tuned as a standard token classifier does not provide a substantial advantage.

Majority-voting ensembles provide a clearer improvement. The best ensemble run, \texttt{ensemble5}, obtains a micro-averaged $F_1$ of 0.8244, corresponding to an absolute improvement of 0.0248 over the baseline and 0.0236 over the best single-model \ac{NER} run. This improvement is mostly driven by precision, which increases from 0.7782 for the baseline to 0.8639 for the best ensemble. This behavior is coherent with the majority-voting strategy, since predictions need to be supported by multiple models in order to be retained, reducing the number of false positives. However, the recall of ensemble systems remains generally below the baseline, showing that this gain in precision is obtained at the cost of discarding some correct mentions.

The best results are obtained by our \ac{EFCL} architecture. All eight submitted \ac{EFCL} runs outperform both the baseline and all ensemble configurations by a large margin. The best run, \texttt{merged6}, which combines \texttt{BiomedBERT-large-uncased-abstract} as term extractor with \texttt{BiomedBERT-large-uncased-abstract} as term classifier, obtains a micro-averaged precision of 0.9548, recall of 0.8058, and $F_1$ of 0.8740. This corresponds to an absolute improvement of 0.0744 $F_1$ points over the baseline and 0.0496 over the best ensemble run. Importantly, the performance gain is again mainly due to Precision: compared with the baseline, the best \ac{EFCL} system improves Precision by 0.1766 points, while Recall decreases only moderately, from 0.8221 to 0.8058.
This result is consistent with the intended precision-oriented behavior behind the \ac{EFCL} architecture, as discussed in Section~\ref{sec:ner-module}. 
Indeed, across all submitted \ac{EFCL} configurations, Precision is always above 0.93.

Considering the official leaderboard with one best run per team, our best submission is also the best-performing \ac{NER} run overall, reaching the highest micro-averaged $F_1$ score. This shows that the proposed \ac{TE}+\ac{TC} strategy is not only more effective than our internal single-model and ensemble baselines, but also competitive with respect to the systems submitted by other participating teams.

\begin{table}[t]
\caption{Micro-averaged \ac{NERD} results on the official test set. Each run combines one of the six best \ac{EFCL} \ac{NER} configurations with one of the four selected \ac{NEL} configurations. The \textit{Run ID} column identifies the \ac{NEL} configuration, while the \textit{NER} column reports the upstream \ac{NER} configuration used to extract entity mentions. The \textit{Retriever} and \textit{Reranker} columns report, respectively, the text-embedding retriever and cross-encoder reranker models used in the \ac{NEL} module. Performance is reported in terms of precision ($P$), recall ($R$), and $F_1$ score, with runs sorted by decreasing $F_1$. The official baseline is highlighted.}
\label{tab:nerd-test}
\begin{tabular}{llllccc}
\toprule
Run ID & NER & Retriever & Reranker & $P$ & $R$ & $F_1$ \\
\midrule
24 & merged6 & BiomedBERT  & BiomedBERT  & 0.7527 & 0.6353 & 0.6890 \\ \midrule
16 & merged4 & BiomedBERT  & BiomedBERT  & 0.7511 & 0.6342 & 0.6877 \\ \midrule
20 & merged5 & BiomedBERT  & BiomedBERT  & 0.7511 & 0.6342 & 0.6877 \\ \midrule
4  & merged1 & BiomedBERT  & BiomedBERT  & 0.7530 & 0.6327 & 0.6876 \\ \midrule
12 & merged3 & BiomedBERT  & BiomedBERT  & 0.7509 & 0.6334 & 0.6872 \\ \midrule
3  & merged1 & BiomedBERT  & BioLinkBERT & 0.7517 & 0.6316 & 0.6864 \\ \midrule
23 & merged6 & BiomedBERT  & BioLinkBERT & 0.7492 & 0.6323 & 0.6858 \\ \midrule
8  & merged2 & BiomedBERT  & BiomedBERT  & 0.7489 & 0.6323 & 0.6857 \\ \midrule
1  & merged1 & BioLinkBERT & BioLinkBERT & 0.7503 & 0.6305 & 0.6852 \\ \midrule
15 & merged4 & BiomedBERT  & BioLinkBERT & 0.7480 & 0.6316 & 0.6849 \\ \midrule
19 & merged5 & BiomedBERT  & BioLinkBERT & 0.7480 & 0.6316 & 0.6849 \\ \midrule
7  & merged2 & BiomedBERT  & BioLinkBERT & 0.7471 & 0.6308 & 0.6841 \\ \midrule
11 & merged3 & BiomedBERT  & BioLinkBERT & 0.7474 & 0.6305 & 0.6840 \\ \midrule
21 & merged6 & BioLinkBERT & BioLinkBERT & 0.7466 & 0.6301 & 0.6834 \\ \midrule
17 & merged5 & BioLinkBERT & BioLinkBERT & 0.7458 & 0.6297 & 0.6829 \\ \midrule
13 & merged4 & BioLinkBERT & BioLinkBERT & 0.7454 & 0.6294 & 0.6825 \\ \midrule
5  & merged2 & BioLinkBERT & BioLinkBERT & 0.7454 & 0.6294 & 0.6825 \\ \midrule
2  & merged1 & BioLinkBERT & BiomedBERT  & 0.7472 & 0.6279 & 0.6824 \\ \midrule
9  & merged3 & BioLinkBERT & BioLinkBERT & 0.7456 & 0.6290 & 0.6823 \\ \midrule
22 & merged6 & BioLinkBERT & BiomedBERT  & 0.7453 & 0.6290 & 0.6822 \\ \midrule
18 & merged5 & BioLinkBERT & BiomedBERT  & 0.7432 & 0.6275 & 0.6805 \\ \midrule
10 & merged3 & BioLinkBERT & BiomedBERT  & 0.7434 & 0.6271 & 0.6803 \\ \midrule
14 & merged4 & BioLinkBERT & BiomedBERT  & 0.7428 & 0.6271 & 0.6801 \\ \midrule
6  & merged2 & BioLinkBERT & BiomedBERT  & 0.7423 & 0.6268 & 0.6797 \\ \midrule
\rowcolor[HTML]{C9DAF8}
BASELINE & GLiNER & \multicolumn{2}{c}{3-stage} & 0.4281 & 0.4522 & 0.4398 \\
\bottomrule
\end{tabular}
\end{table}

\subsection{Named Entity Recognition and Disambiguation (Subtask 6.1.2)}
\label{sec:nerd-test-results}

For the \ac{NERD} subtask, we submitted 24 runs obtained by combining the six best \ac{NER} configurations selected on the development set with the four \ac{NEL} configurations discussed in Section~\ref{sec:nerd-dev-results}. All six upstream \ac{NER} systems are based on our two-stage \ac{EFCL} architecture. For \ac{NEL}, we fixed the architecture to a text embedding retriever followed by a cross-encoder reranker, varying the transformer backbone used by the two components between \texttt{BiomedBERT} and \texttt{BioLinkBERT}, as explained in Section~\ref{sec:nerd-dev-results}. Table~\ref{tab:nerd-test} reports the micro-averaged results obtained by our submitted runs on the official test set, together with the official baseline.

The results show a large improvement over the baseline across all submitted runs. The best run, \texttt{24merged6}, combines the best upstream \ac{EFCL} configuration with a \texttt{BiomedBERT}-based text embedding retriever and a \texttt{BiomedBERT}-based cross-encoder reranker. It obtains a micro-averaged Precision of 0.7527, Recall of 0.6353, and $F_1$ of 0.6890. Compared with the official baseline, which obtains 0.4398 $F_1$, this corresponds to an absolute improvement of 0.2492 $F_1$ points. All submitted runs remain well above the baseline, with scores ranging from 0.6797 to 0.6890.

As observed on the development set in Section~\ref{sec:nerd-dev-results}, the four \ac{NEL} configurations lead to very similar results. The performance range among our 24 runs is narrow, with less than one $F_1$ point between the best and the lowest-scoring submission. This confirms that, once the retriever--reranker architecture is fixed, changing the transformer backbone between \texttt{BiomedBERT} and \texttt{BioLinkBERT} has only a limited impact. The best-performing configurations tend to use \texttt{BiomedBERT} in both the retriever and reranker, but the difference with the other combinations remains small.

These results should be interpreted by considering the strong dependency of \ac{NERD} on the upstream \ac{NER} module. Since our \ac{NEL} component assigns exactly one concept to every entity mention it receives, it cannot recover entity mentions missed by \ac{NER}, neither it can remove incorrectly extracted spans. Therefore, the final \ac{NERD} score reflects both mention extraction quality and concept assignment quality. For this reason, while our best run also ranks first in the official \ac{NERD} leaderboard when considering the best submission per team, we cannot conclude from the final \ac{NERD} score alone that our \ac{NEL} module is better than those of other participants. Nevertheless, the fact that our margin over the second-best team is larger in \ac{NERD} than in \ac{NER} suggests that the proposed linking module preserves and possibly amplifies the advantage obtained by the upstream \ac{EFCL} architecture.

Overall, the test-set results support the design decisions made on the development set. The two-stage \ac{TE}+\ac{TC} architecture provides high-quality entity mentions, while the retriever-reranker \ac{NEL} module assigns concepts with stable performance across model combinations. The resulting end-to-end \ac{NERD} pipeline substantially outperforms the official baseline and achieves the best score among participating systems.

\begin{table}[t]
\caption{Micro-averaged Mention-Level RE results on the official test set. Column ``NER" reports the Run ID of the configuration used to predict the entity annotations (see Table~\ref{tab:ner-test} for details on the configurations). Row ``Pre-Train" reports the pre-training configuration. For instance, ``sb10" refers to ATLOP pre-trained for 10 epochs on the silver and bronze annotations. Row ``Finetune" reports the finetuning configuration. For instance, ``g100" refers to ATLOP finetuned for 100 epochs on the gold annotations. Runs are listed on descending order of $F_1$.}
\label{tab:mre-test}
\begin{tabular}{llllccc}
\toprule
Run ID   & NER & Pre-Train & Fine-Tune & $P$    & $R$    & $F_1$  \\ \midrule
mre10    & merged4 & sb10 & g100  & 0.8996 & 0.4651 & 0.6132 \\ \hline
mre11    & merged5 & sb10 & g100 & 0.8996 & 0.4651 & 0.6132 \\ \hline
mre8     & merged2 & sb10 & g100 & 0.8996 & 0.4651 & 0.6132 \\ \hline
mre7     & merged1 & sb10 & g100 & 0.8974 & 0.4651 & 0.6127 \\ \hline
mre9     & merged3 & sb10 & g100 & 0.8821 & 0.4651 & 0.6091 \\ \hline
mre12    & merged6 & sb10 & g100 & 0.8811 & 0.4651 & 0.6088 \\ \hline
mre13    & merged1 & sb5 & g50 & 0.8726 & 0.4651 & 0.6068 \\ \hline
mre14    & merged2 & sb5 & g50 & 0.8715 & 0.4651 & 0.6065 \\ \hline
mre17    & merged5 & sb5 & g50 & 0.8715 & 0.4651 & 0.6065 \\ \hline
mre16    & merged4 & sb5 & g50 & 0.8695 & 0.4651 & 0.6060 \\ \hline
mre20    & merged2 & sb2 & g10 & 0.8612 & 0.4651 & 0.6040 \\ \hline
mre22    & merged4 & sb2 & g10 & 0.8612 & 0.4651 & 0.6040 \\ \hline
mre23    & merged5 & sb2 & g10 & 0.8612 & 0.4651 & 0.6040 \\ \hline
mre19    & merged1 & sb2 & g10 & 0.8582 & 0.4651 & 0.6032 \\ \hline
mre15    & merged3 & sb5 & g50 & 0.8511 & 0.4651 & 0.6015 \\ \hline
mre21    & merged3 & sb2 & g10 & 0.8511 & 0.4651 & 0.6015 \\ \hline
mre24    & merged6 & sb2 & g10 & 0.8481 & 0.4651 & 0.6007 \\ \hline
mre18    & merged6 & sb5 & g50 & 0.8471 & 0.4651 & 0.6005 \\ \hline
mre5     & merged5 & b2 & gs10 & 0.8374 & 0.4651 & 0.5980 \\ \hline
mre1     & merged1 & b2 & gs10 & 0.8354 & 0.4651 & 0.5975 \\ \hline
mre2     & merged2 & b2 & gs10 & 0.8354 & 0.4651 & 0.5975 \\ \hline
mre4     & merged4 & b2 & gs10 & 0.8335 & 0.4651 & 0.5970 \\ \hline
mre3     & merged3 & b2 & gs10 & 0.8259 & 0.4651 & 0.5951 \\ \hline
mre6     & merged6 & b2 & gs10 & 0.8222 & 0.4651 & 0.5941 \\ \midrule
\rowcolor[HTML]{C9DAF8} 
BASELINE & GLiNER & --- & ATLOP & 0.4444 & 0.3453 & 0.3886 \\ \bottomrule
\end{tabular}%
\end{table}

\subsection{Mention-Level RE (Subtask 6.2.1)}

For mention-level RE, we submitted 24 runs obtained by combining the six best \ac{NERD} configurations selected based on the best micro-F1 on the development dataset (see Table~\ref{tab:termMerged-dev-top15}), and the four best mention-level RE based on the highest micro-F1 on the developmenet dataset (see Table~\ref{tab:mre-dev}). Table~\ref{tab:mre-test} reports the micro-averaged performance of our model on the mention-level RE subtask. Overall, all our runs show a significant improvement with respect to the baseline, especially in precision. Compared to the baseline, our best run doubles in precision and reports an improvement of 11 points in recall and 22 points in F1. On average, our model yields a precision gain of $+93.5\%$, $+34.7\%$ in recall, and $+55.3\%$ F1. The test-set results corroborate the findings observed on the development set: our model is precision-oriented, increasing the number of epochs produces more precise predictions. As for the development dataset, the best performing RE model is pre-trained on the silver and bronze annotations for 10 epochs and finetuned on the gold dataset for 100 epochs. Under this configuration, three NER models produce identical outcomes; all three are within the top-4 for the NER subtask. The results on the test dataset are not deteriorated with respect to those on the development dataset, demonstrating that our end-to-end approach leverages strong NER models that do not propagate noise to the RE module.

\begin{table}[b]
\caption{Micro-averaged Concept-Level RE results on the official test set. Column ``NER+RE" reports the Run ID of the configuration used to predict the entity annotations and perform relation extraction (see Table~\ref{tab:mre-test} for details on the configurations). Row ``Retriever" reports the retriever configuration. For instance, ``textemb-BiomedBERT" refers to the text embedding retriever using retriever model BiomedBERT. Row ``Reranker" reports the reranker configuration. For instance, ``crossenc-BioLinkBERT" refers to the cross encoder reranker using BioLinkBERT as reranker model. Runs are listed on descending order of $F_1$.}
\label{tab:cre-test}
\resizebox{\textwidth}{!}{%
\begin{tabular}{llllccc}
\toprule
Run ID   & NER+RE & Retriever & Reranker & $P$    & $R$    & $F_1$  \\ \midrule
11mre12  & mre12 & textemb-BiomedBERT & crossenc-BioLinkBERT  & 0.4903 & 0.2691 & 0.3475 \\ \hline
15mre7   & mre7 & textemb-BiomedBERT & crossenc-BioLinkBERT  & 0.4903 & 0.2691 & 0.3475 \\ \hline
19mre8   & mre8 & textemb-BiomedBERT & crossenc-BioLinkBERT  & 0.4903 & 0.2691 & 0.3475 \\ \hline
23mre9   & mre9 & textemb-BiomedBERT & crossenc-BioLinkBERT  & 0.4903 & 0.2691 & 0.3475 \\ \hline
12mre12  & mre12 & textemb-BiomedBERT & crossenc-BiomedBERT   & 0.4895 & 0.2691 & 0.3473 \\ \hline
16mre7   & mre7 & textemb-BiomedBERT & crossenc-BiomedBERT   & 0.4895 & 0.2691 & 0.3473 \\ \hline
20mre8   & mre8 & textemb-BiomedBERT & crossenc-BiomedBERT   & 0.4895 & 0.2691 & 0.3473 \\ \hline
24mre9   & mre9 & textemb-BiomedBERT & crossenc-BiomedBERT   & 0.4895 & 0.2691 & 0.3473 \\ \hline
3mre10   & mre10 & textemb-BiomedBERT & crossenc-BioLinkBERT  & 0.4888 & 0.2691 & 0.3471 \\ \hline
7mre11   & mre11 & textemb-BiomedBERT & crossenc-BioLinkBERT  & 0.4888 & 0.2691 & 0.3471 \\ \hline
4mre10   & mre10 & textemb-BiomedBERT & crossenc-BiomedBERT   & 0.4881 & 0.2691 & 0.3469 \\ \hline
8mre11   & mre11 & textemb-BiomedBERT & crossenc-BiomedBERT   & 0.4881 & 0.2691 & 0.3469 \\ \hline
10mre12  & mre12 & textemb-BioLinkBERT & crossenc-BiomedBERT  & 0.4814 & 0.2658 & 0.3425 \\ \hline
14mre7   & mre7 & textemb-BioLinkBERT & crossenc-BiomedBERT  & 0.4814 & 0.2658 & 0.3425 \\ \hline
18mre8   & mre8 & textemb-BioLinkBERT & crossenc-BiomedBERT  & 0.4814 & 0.2658 & 0.3425 \\ \hline
22mre9   & mre9 & textemb-BioLinkBERT & crossenc-BiomedBERT  & 0.4814 & 0.2658 & 0.3425 \\ \hline
2mre10   & mre10 & textemb-BioLinkBERT & crossenc-BiomedBERT  & 0.4799 & 0.2658 & 0.3422 \\ \hline
6mre11   & mre11 & textemb-BioLinkBERT & crossenc-BiomedBERT  & 0.4799 & 0.2658 & 0.3422 \\ \hline
13mre7   & mre7 & textemb-BioLinkBERT & crossenc-BioLinkBERT & 0.4717 & 0.2609 & 0.3360 \\ \hline
17mre8   & mre8 & textemb-BioLinkBERT & crossenc-BioLinkBERT & 0.4717 & 0.2609 & 0.3360 \\ \hline
21mre9   & mre9 & textemb-BioLinkBERT & crossenc-BioLinkBERT & 0.4717 & 0.2609 & 0.3360 \\ \hline
9mre12   & mre12 & textemb-BioLinkBERT & rossenc-BioLinkBERT & 0.4717 & 0.2609 & 0.3360 \\ \hline
1mre10   & mre10 & textemb-BioLinkBERT & crossenc-BioLinkBERT & 0.4703 & 0.2609 & 0.3356 \\ \hline
5mre11   & mre11 & textemb-BioLinkBERT & crossenc-BioLinkBERT & 0.4703 & 0.2609 & 0.3356 \\ \midrule
\rowcolor[HTML]{C9DAF8} 
BASELINE & GLiNER + ATLOP & \multicolumn{2}{c}{3-stage} & 0.1403 & 0.1292 & 0.1345 \\ \bottomrule
\end{tabular}%
}
\end{table}

\subsection{Concept-Level RE (Subtask 6.2.2)}
For \ac{C-RE}, we submitted 24 runs obtained by combining the six best \ac{M-RE} configurations, selected according to micro-averaged $F_1$ on the development set (see Table~\ref{tab:mre-dev}), with the four \ac{NEL} configurations discussed in Section~\ref{sec:nerd-dev-results}, which are the same used in the runs submitted for the \ac{NERD} subtask (see Section~\ref{sec:nerd-test-results}). The selected \ac{M-RE} configurations were all trained using either the 10-epoch pre-training and 100-epoch fine-tuning setting, or the 5-epoch pre-training and 50-epoch fine-tuning setting. 

Table~\ref{tab:cre-test} reports the micro-averaged performance of our model on the \ac{C-RE} subtask. All our runs report a significant improvement with respect to the baseline. The precision ranges from $47.03\%$ to $49.03\%$, more than three times higher than the baseline precision, registering an average improvement of $+244\%$ compared to the baseline. The recall ranges from $26.91\%$ to $26.09\%$, more than double the baseline recall, reporting an average percentage gain of $+106\%$. The F1 score ranges from $34.75\%$ to $33.56\%$, 2.5 times higher than the baseline F1, corresponding to an average increase of $+155\%$. 



\section{Conclusion}
\label{sec:conclusion}
We proposed \ac{TWIX}, a precision-oriented \ac{IE} pipeline to solve the four GutBrainIE subtasks. \ac{TWIX} consists of three interconnected modules, addressing \ac{NER}, \ac{NEL}, and \ac{RE}. The \ac{NER} module features an \ac{EFCL} architecture, designed to first identify candidate entity mentions and assign an entity type to each mention in a subsequent step. The \ac{NEL} module follows a retrieve-and-rerank strategy: a retriever first selects the top-$k$ candidate concepts for each mention, and a reranker then selects the most likely concept in context. The \ac{RE} module builds upon ATLOP and leverages a two-stage training framework, pre-training the model on noisy labels and finetuning it on high-quality annotations. Results on the test dataset shows that our architecture systematically improves the baseline performance. The top-5 performing \ac{NER} models confirms the \ac{EFCL} architecture is effective for the subtask; precision is improved by eight points compared to the baseline ($87.40\%$ vs $79.96\%$). The \ac{NERD} module demonstrates to be effective by almost doubling the baseline performance in all configurations. The \ac{RE} module shows a significant improvement both in mention-level RE (F1 from $38.86\%$ to $61.32\%$) and concept-level RE (F1 increases from $13.45\%$ to $34.75\%$).

\begin{acknowledgments}
This work has received funding from the HEREDITARY Project as part of the European Union’s Horizon Europe research and innovation programme under grant agreement No GA 101137074. 
Views and opinions expressed are, however, those of the authors only and do not necessarily reflect those of the European Union. Neither the European Union nor the granting authority can be held responsible. 
\end{acknowledgments}

\section*{Declaration on Generative AI}
During the preparation of this work, the authors used ChatGPT, Grammarly, and Comet in order to: Improve writing style, and Formatting assistance. After using these tools/services, the authors reviewed and edited the content as needed and take full responsibility for the publication’s content.

\bibliography{references}

\clearpage
\newpage
\appendix

\section{\ac{EFCL} \ac{NER} Results on the Development Set}
Table \ref{tab:termMerged-dev-full} reports \ac{NER} results for all the 49 \ac{EFCL} configurations we tested on the development set. 

\begin{table}[b]
\captionsetup{font=small} 
\caption{Micro-averaged \ac{NER} performance on the development set for the proposed \ac{EFCL} architecture. Each row reports a combination of \ac{TE} and \ac{TC} models, with runs sorted by decreasing $F_1$ score. Performance is reported in terms of precision ($P$), recall ($R$), and $F_1$ score. Best results for each metric are bolded and second-best results are underlined.}
\label{tab:termMerged-dev-full}
\resizebox{0.9\textwidth}{!}{%
\begin{tabular}{lllccc}
\toprule
ID & Term Extractor Model                                & Term Classifier Model                               & $P$    & $R$    & $F_1$  \\ \midrule
1  & BiomedNLP-BiomedBERT-large-uncased-abstract         & BiomedNLP-BiomedBERT-large-uncased-abstract         & 0.9544 & \textbf{0.8382} & \textbf{0.8925} \\ \midrule
2  & BiomedNLP-BiomedBERT-large-uncased-abstract         & BiomedNLP-BiomedBERT-base-uncased-abstract-fulltext & 0.9544 & \textbf{0.8382} & \textbf{0.8925} \\ \midrule
3  & BiomedNLP-BiomedBERT-large-uncased-abstract         & BiomedNLP-BiomedBERT-base-uncased-abstract          & 0.9507 & \underline{0.8346} & \underline{0.8889} \\ \midrule
4  & BiomedNLP-BiomedBERT-large-uncased-abstract         & BioLinkBERT-large                                   & 0.9494 & 0.8338 & 0.8879 \\ \midrule
5  & BiomedNLP-BiomedBERT-large-uncased-abstract         & BioLinkBERT-base                                    & 0.9485 & 0.8326 & 0.8868 \\ \midrule
6  & BiomedNLP-BiomedBERT-large-uncased-abstract         & BiomedNLP-BiomedElectra-base-uncased-abstract       & 0.9476 & 0.8326 & 0.8864 \\ \midrule
7  & BioLinkBERT-base                                    & BiomedNLP-BiomedBERT-large-uncased-abstract         & 0.9472 & 0.8322 & 0.8860 \\ \midrule
8  & BiomedNLP-BiomedElectra-large-uncased-abstract      & BiomedNLP-BiomedBERT-large-uncased-abstract         & 0.9516 & 0.8259 & 0.8843 \\ \midrule
9  & BiomedNLP-BiomedBERT-large-uncased-abstract         & BiomedNLP-BiomedElectra-large-uncased-abstract      & \textbf{0.9574} & 0.8207 & 0.8838 \\ \midrule
10 & BioLinkBERT-base                                    & BiomedNLP-BiomedBERT-base-uncased-abstract-fulltext & 0.9511 & 0.8255 & 0.8838 \\ \midrule
11 & BiomedNLP-BiomedBERT-base-uncased-abstract-fulltext & BiomedNLP-BiomedBERT-large-uncased-abstract         & 0.9535 & 0.8219 & 0.8828 \\ \midrule
12 & BiomedNLP-BiomedElectra-large-uncased-abstract      & BiomedNLP-BiomedBERT-base-uncased-abstract-fulltext & 0.9561 & 0.8199 & 0.8828 \\ \midrule
13 & BiomedNLP-BiomedBERT-base-uncased-abstract-fulltext & BiomedNLP-BiomedBERT-base-uncased-abstract-fulltext & 0.9531 & 0.8215 & 0.8824 \\ \midrule
14 & BioLinkBERT-large                                   & BiomedNLP-BiomedBERT-base-uncased-abstract-fulltext & \underline{0.9572} & 0.8159 & 0.8809 \\ \midrule
15 & BioLinkBERT-base                                    & BiomedNLP-BiomedElectra-large-uncased-abstract      & 0.9479 & 0.8227 & 0.8809 \\ \midrule
16 & BioLinkBERT-base                                    & BiomedNLP-BiomedBERT-base-uncased-abstract          & 0.9466 & 0.8219 & 0.8798 \\ \midrule
17 & BiomedNLP-BiomedBERT-base-uncased-abstract-fulltext & BiomedNLP-BiomedElectra-large-uncased-abstract      & 0.9503 & 0.8191 & 0.8798 \\ \midrule
18 & BioLinkBERT-large                                   & BiomedNLP-BiomedBERT-large-uncased-abstract         & 0.9558 & 0.8148 & 0.8797 \\ \midrule
19 & BioLinkBERT-base                                    & BioLinkBERT-large                                   & 0.9465 & 0.8215 & 0.8796 \\ \midrule
20 & BiomedNLP-BiomedElectra-large-uncased-abstract      & BiomedNLP-BiomedElectra-large-uncased-abstract      & 0.9523 & 0.8159 & 0.8789 \\ \midrule
21 & BiomedNLP-BiomedBERT-base-uncased-abstract-fulltext & BiomedNLP-BiomedBERT-base-uncased-abstract          & 0.9489 & 0.8179 & 0.8786 \\ \midrule
22 & BiomedNLP-BiomedElectra-large-uncased-abstract      & BiomedNLP-BiomedBERT-base-uncased-abstract          & 0.9514 & 0.8159 & 0.8785 \\ \midrule
23 & BioLinkBERT-base                                    & BioLinkBERT-base                                    & 0.9447 & 0.8203 & 0.8781 \\ \midrule
24 & BiomedNLP-BiomedElectra-large-uncased-abstract      & BioLinkBERT-base                                    & 0.9510 & 0.8155 & 0.8781 \\ \midrule
25 & BiomedNLP-BiomedBERT-base-uncased-abstract-fulltext & BioLinkBERT-large                                   & 0.9485 & 0.8175 & 0.8781 \\ \midrule
26 & BioLinkBERT-base                                    & BiomedNLP-BiomedElectra-base-uncased-abstract       & 0.9443 & 0.8199 & 0.8777 \\ \midrule
27 & BiomedNLP-BiomedElectra-large-uncased-abstract      & BioLinkBERT-large                                   & 0.9500 & 0.8148 & 0.8772 \\ \midrule
28 & BiomedNLP-BiomedBERT-base-uncased-abstract-fulltext & BiomedNLP-BiomedElectra-base-uncased-abstract       & 0.9471 & 0.8167 & 0.8771 \\ \midrule
29 & BiomedNLP-BiomedBERT-base-uncased-abstract-fulltext & BioLinkBERT-base                                    & 0.9471 & 0.8163 & 0.8769 \\ \midrule
30 & BiomedNLP-BiomedElectra-base-uncased-abstract       & BiomedNLP-BiomedBERT-large-uncased-abstract         & 0.9546 & 0.8096 & 0.8762 \\ \midrule
31 & BiomedNLP-BiomedElectra-base-uncased-abstract       & BiomedNLP-BiomedBERT-base-uncased-abstract-fulltext & 0.9546 & 0.8092 & 0.8759 \\ \midrule
32 & BioLinkBERT-large                                   & BiomedNLP-BiomedElectra-large-uncased-abstract      & 0.9516 & 0.8112 & 0.8758 \\ \midrule
33 & BiomedNLP-BiomedElectra-large-uncased-abstract      & BiomedNLP-BiomedElectra-base-uncased-abstract       & 0.9482 & 0.8136 & 0.8757 \\ \midrule
34 & BioLinkBERT-large                                   & BiomedNLP-BiomedBERT-base-uncased-abstract          & 0.9507 & 0.8108 & 0.8752 \\ \midrule
35 & BioLinkBERT-large                                   & BioLinkBERT-base                                    & 0.9502 & 0.8104 & 0.8748 \\ \midrule
36 & BioLinkBERT-large                                   & BioLinkBERT-large                                   & 0.9493 & 0.8096 & 0.8739 \\ \midrule
37 & BiomedNLP-BiomedElectra-base-uncased-abstract       & BiomedNLP-BiomedElectra-large-uncased-abstract      & 0.9514 & 0.8068 & 0.8731 \\ \midrule
38 & BioLinkBERT-large                                   & BiomedNLP-BiomedElectra-base-uncased-abstract       & 0.9479 & 0.8084 & 0.8726 \\ \midrule
39 & BiomedNLP-BiomedElectra-base-uncased-abstract       & BiomedNLP-BiomedBERT-base-uncased-abstract          & 0.9504 & 0.8064 & 0.8725 \\ \midrule
40 & BiomedNLP-BiomedElectra-base-uncased-abstract       & BioLinkBERT-large                                   & 0.9481 & 0.8044 & 0.8704 \\ \midrule
41 & BiomedNLP-BiomedElectra-base-uncased-abstract       & BiomedNLP-BiomedElectra-base-uncased-abstract       & 0.9481 & 0.8044 & 0.8704 \\ \midrule
42 & BiomedNLP-BiomedElectra-base-uncased-abstract       & BioLinkBERT-base                                    & 0.9472 & 0.8036 & 0.8695 \\ \midrule
43 & BiomedNLP-BiomedBERT-base-uncased-abstract          & BiomedNLP-BiomedBERT-base-uncased-abstract-fulltext & 0.9569 & 0.7925 & 0.8670 \\ \midrule
44 & BiomedNLP-BiomedBERT-base-uncased-abstract          & BiomedNLP-BiomedBERT-large-uncased-abstract         & 0.9550 & 0.7910 & 0.8653 \\ \midrule
45 & BiomedNLP-BiomedBERT-base-uncased-abstract          & BiomedNLP-BiomedElectra-large-uncased-abstract      & 0.9511 & 0.7878 & 0.8618 \\ \midrule
46 & BiomedNLP-BiomedBERT-base-uncased-abstract          & BiomedNLP-BiomedBERT-base-uncased-abstract          & 0.9502 & 0.7874 & 0.8612 \\ \midrule
47 & BiomedNLP-BiomedBERT-base-uncased-abstract          & BioLinkBERT-base                                    & 0.9497 & 0.7870 & 0.8607 \\ \midrule
48 & BiomedNLP-BiomedBERT-base-uncased-abstract          & BioLinkBERT-large                                   & 0.9478 & 0.7854 & 0.8590 \\ \midrule
49 & BiomedNLP-BiomedBERT-base-uncased-abstract          & BiomedNLP-BiomedElectra-base-uncased-abstract       & 0.9469 & 0.7846 & 0.8581 \\ 
\bottomrule
\end{tabular}%
}
\end{table}

\end{document}